\documentclass[10pt,twocolumn,letterpaper]{article}

\usepackage{cvpr}
\usepackage{times}
\usepackage{epsfig}
\usepackage{graphicx}
\usepackage{amsmath}
\usepackage{amssymb}
\usepackage{xcolor}
\usepackage{caption}
\usepackage{algorithm}
\usepackage{algorithmic}
\usepackage{subcaption}
\usepackage{wrapfig}
\usepackage[pagebackref=true,breaklinks=true,letterpaper=true,colorlinks,bookmarks=false]{hyperref}
\usepackage{kky}

 \cvprfinalcopy 

\def\cvprPaperID{6834} 

\ifcvprfinal\fi
\begin{document}
\setlength{\abovedisplayskip}{6pt}
\setlength{\belowdisplayskip}{6pt}


\title{LBS Autoencoder: Self-supervised Fitting of Articulated Meshes to Point Clouds}

\author{Chun-Liang Li$^1$, Tomas Simon$^2$, Jason Saragih$^2$, Barnab\'as P\'oczos$^1$, Yaser Sheikh$^{1,2}$\\
 $^1$Carnegie Mellon University and $^2$Facebook Reality Labs\\
{\tt\small \{chunlial, bapoczos\}@cs.cmu.edu} \hspace{1em} {\tt\small \{firstname.lastname\}@fb.com}
}

\twocolumn[{%
\renewcommand\twocolumn[1][]{#1}%
\maketitle
}]
\begin{abstract}
We present LBS-AE; a self-supervised autoencoding algorithm for fitting articulated mesh models to point clouds.
As input, we take a sequence of point clouds to be registered as well as an artist-rigged mesh, \ie a template mesh equipped with a linear-blend skinning (LBS) deformation space parameterized by a skeleton hierarchy. As output, we learn an LBS-based autoencoder that produces registered meshes from the input point clouds. 
To bridge the gap between the artist-defined geometry and the captured point clouds, our autoencoder models pose-dependent deviations from the template geometry. 
During training, instead of using explicit correspondences, such as key points or pose supervision, our method leverages LBS deformations to bootstrap the learning process. To avoid poor local minima from erroneous point-to-point correspondences, we utilize a {\em structured} Chamfer distance based on part-segmentations, which are learned concurrently using self-supervision.  We demonstrate qualitative results on real captured hands, and report quantitative evaluations on the FAUST benchmark for body registration. Our method achieves performance that is superior to other unsupervised approaches and comparable to methods using supervised examples.

\end{abstract}
\begin{figure}
    \centering
    \includegraphics[width=1.0\linewidth]{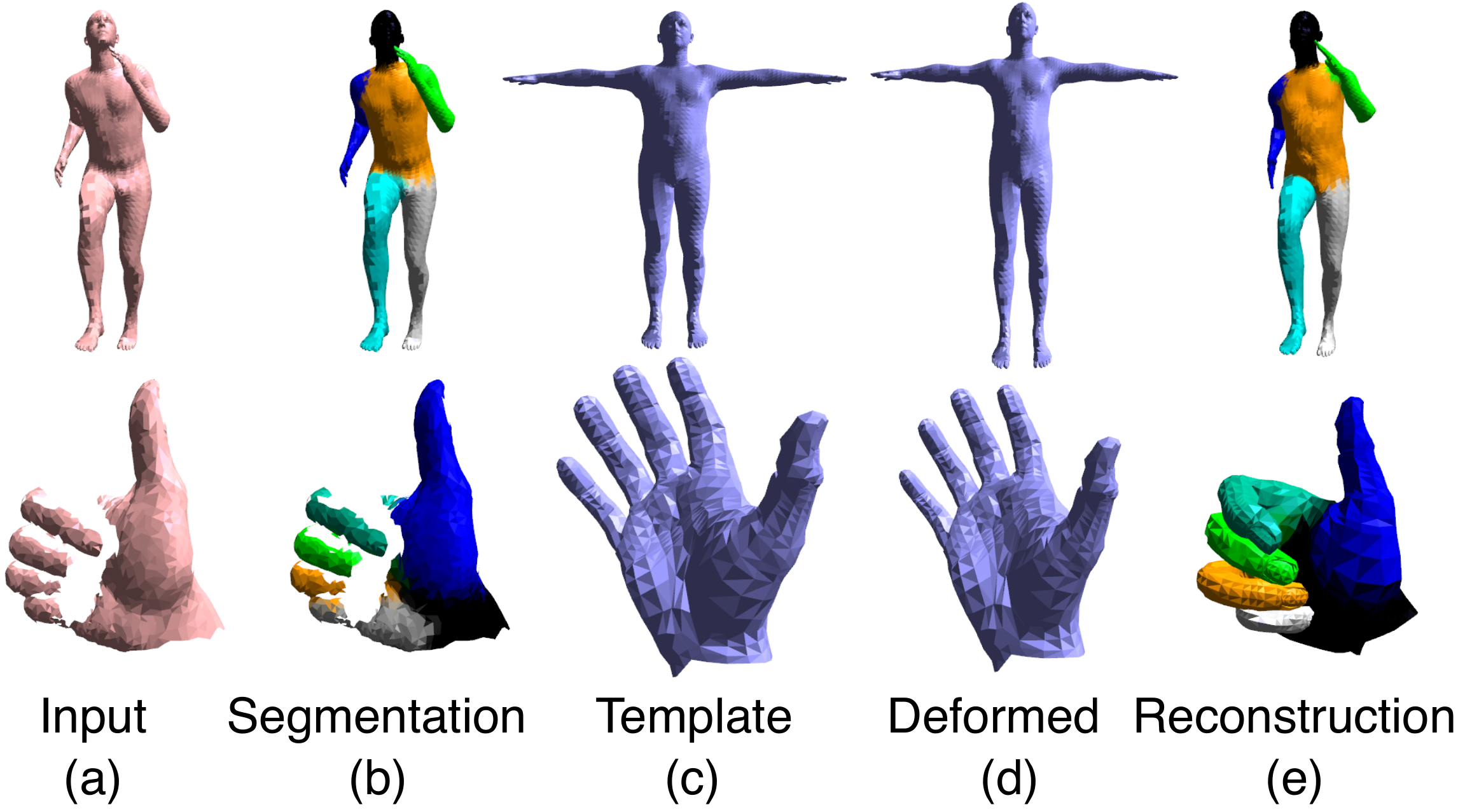}
	\vspace{-20pt}
    \captionof{figure}{
		Given point clouds sampled from the surface of an input shape (a), our model infers a coarse segmentation (b), and learns to deform a given template (c), through a combination of deformations of the template (d) as well as pose deformation parameterized by LBS to match the reconstruction (e). 
		We use a \emph{structured Chamfer distance} that uses the inferred segmentation of the data (b) as coarse correspondence to measure distance between matching regions to avoid local optima in the Chamfer distance.
	}
	\vspace{-10pt}
    \label{fig:teaser}
\end{figure}%
\section{Introduction}

The registration of unstructured point-clouds to a common mesh representation is an important problem in computer vision and has been extensively studied in the past decades.
Works in this area can be coarsely grouped together based on how much prior knowledge and supervision is incorporated into the fitting method. On one end of the spectrum, there are entirely unsupervised and object-agnostic models, such as FoldingNet \cite{yang2018foldingnet} or AtlasNet~\cite{groueix2018atlasnet}. These methods learn to deform a flat 2D surface to match the target geometry, while making no assumptions about the objects being modeled other than that they can be represented as a 2D surface. Adding slightly more prior knowledge, 3D-CODED~\cite{groueix20183d} uses a template mesh (\eg hand or body) with a topology better suited to the object of interest.

On the other end of the spectrum are highly specialized models for specific objects, such as hands and bodies. Works of
this kind include SCAPE~\cite{Anguelov2005}, Dyna~\cite{pons2015dyna}, SMPL~\cite{loper2015smpl}, and
MANO~\cite{romero2017embodied}. These models are built using high-resolution 3D scans with correspondence and human
curation. They model correctives for different poses and modalities (\eg body types) and can be used as high-quality
generative models of geometry. A number of works learn to manipulate these models to fit data based on different sources
of supervision, such as key points~\cite{bogo2016keep,lassner2017unite,mehta2017vnect,tung2017self,joo2018total} and/or prior distributions of model parameters~\cite{kanazawa2018learning, kanazawa2018end}.

In this paper, we present an unsupervised/self-supervised algorithm, LBS Autoencoder (LBS-AE), to fit such articulated mesh models to point 
cloud data. The proposed algorithm is a middle ground of the two ends of spectrum discussed above in two senses. 

First, 
we assume an articulated template model of the object class is available, but not the statistics of its articulation in our dataset nor the specific shape of the object instance. We argue that this prior information is widely available for many common objects of interest in the form of ``rigged'' or ``skinned'' mesh models, which are typically created by artists for use in animation. In addition to a template mesh describing the geometric shape, these prior models have two more components: (1) a kinematic
hierarchy of transforms describing the degrees of freedom, and (2) a skinning function that defines how transforms in the hierarchy influence each of the mesh vertices. This enables registration to data by manipulating the transforms in the model.
One common example is Linear Blending Skinning (LBS). Therefore,
instead of relying on deep networks to learn the full deformation process from a single template~\cite{groueix20183d}, 
we leverage LBS as part of the decoder to 
model coarse joint deformations. 
Different from hand-crafted models such as SMPL~\cite{loper2015smpl}, LBS by itself does not model pose-dependent correctives between the template and data, nor does it model the space of non-articulated shape variation (\eg body shape). To model these, we also allow our network to learn deformations of the template mesh which, when posed by LBS, result in a better fit to the data. The encoder therefore learns a latent representation from which it can infer both joint angles for use by the LBS deformation, as well as corrective deformations to the template mesh.

Second, for fitting models to data during the training, existing works either rely on explicit supervision (\eg
correspondence~\cite{groueix20183d} and key points~\cite{joo2018total}) or unsupervised nearest neighbors search (\eg
Chamfer Distance (CD)~\cite{yang2018foldingnet}) to find point correspondence between the model and data for measuring reconstruction loss. Rather than using external supervision, we introduce a ``Structured Chamfer Distance'' (SCD), which
improves the blind nearest neighbor search in CD based on an inferred coarse correspondence.  The idea is to
segment the point clouds into corresponding regions (we use regions defined by the LBS weighting). After inferring the segmentation on the input point cloud and the template,
we then apply nearest neighbor search between corresponding regions as high-level correspondence.  The
challenge is we do not assume external supervision to be available for the input point clouds. Instead, we utilize the learned
LBS-AE model to generate \emph{self-supervision} to train the segmentation network from scratch. As the LBS-AE fitting is improved
during training, the training data from self-supervision for segmentation also improves, leading to improved segmentation of the real data.  We are then able to use the improved segmentation to achieve better correspondence and in turn better LBS-AE model fitting. In this paper, we present a joint training framework to learn these two components simultaneously.  Since LBS-AE does not require
any explicit correspondence nor key points, it is similar to approaches which are sometimes referred to as  ``unsupervised'' in the pose
estimation literature~\cite{tewari2018high, genova2018unsupervised}, but it is different from existing unsupervised
learning approach~\cite{yang2018foldingnet} in that it leverages LBS deformation to generate \emph{self-supervision} during training.

In this work, we show that the space of deformations described by an artist-defined rig may sometimes already
be sufficiently constrained to allow fitting to real data without any additional labeling.  Such a model-fitting pipeline without additional supervision has the potential to
simplify geometric registration tasks by requiring less human labeling effort. For example, when fitting an
artist-defined hand rig to point clouds of hands, our method allows for unsupervised hand pose estimation. When fitting a body model to 3D scans of body data, this allows recovering the joint angles of the body as well as registering the
mesh vertices.  In the experiments, we present the results on fitting real hands as well as benchmark body data on the SURREAL and FAUST datasets.

\section{Proposed Method}

We propose to learn a function $\Fcal(\cdot)$ that takes as input an unstructured point cloud $\Xb{=}\{x\}_{i=1}^n$,
where each $x_i$ is a 3D point and $n$ is a variable number, and produces as output a fixed number $m$ of corresponded
vertices $\Vb{=}\{v_i\}_{i=1}^{m}$, where
$\Vb = \Fcal\left( \Xb\right)$.
The vertices $\Vb$ form a mesh with fixed topology whose geometry should closely match that of the input\footnote{Note
that, although we assume the inputs are point clouds, they could also be the vertices of a mesh without using any
topology information.}. Rather than allowing
$\Fcal(\cdot)$ to be any arbitrary deformation produced by a deep neural network (as in~\cite{yang2018foldingnet,groueix2018atlasnet}), 
we force the output to be produced by Linear Blending 
Skinning (LBS) to explicitly encode the motion of joints. We allow additional non-linear deformations (also given by a neural network) to model deviations from the LBS approximation. However, an important difference with respect to similar models, such as SMPL~\cite{loper2015smpl} or MANO~\cite{romero2017embodied}, is that we do not pre-learn the space of non-LBS deformations on a curated set (and then fix them) but rather learn these simultaneously on the data that is to be aligned, with no additional supervision.

\begin{figure}
    \centering
    \begin{subfigure}[b]{0.15\textwidth}
        \centering
        \includegraphics[width=\textwidth]{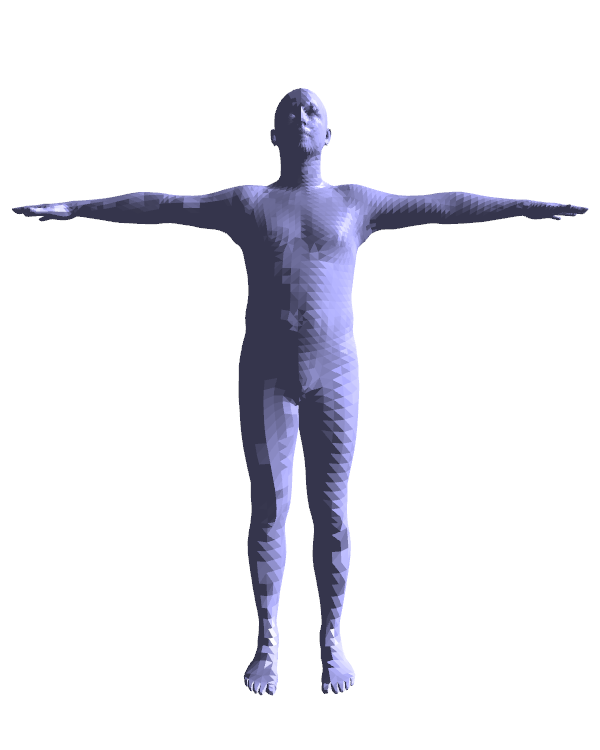}
        \vspace{-20pt}
        \caption{$\Ub$}
		\label{fig:lbs_temp}
    \end{subfigure}
    \begin{subfigure}[b]{0.15\textwidth}
        \centering
        \includegraphics[width=\textwidth]{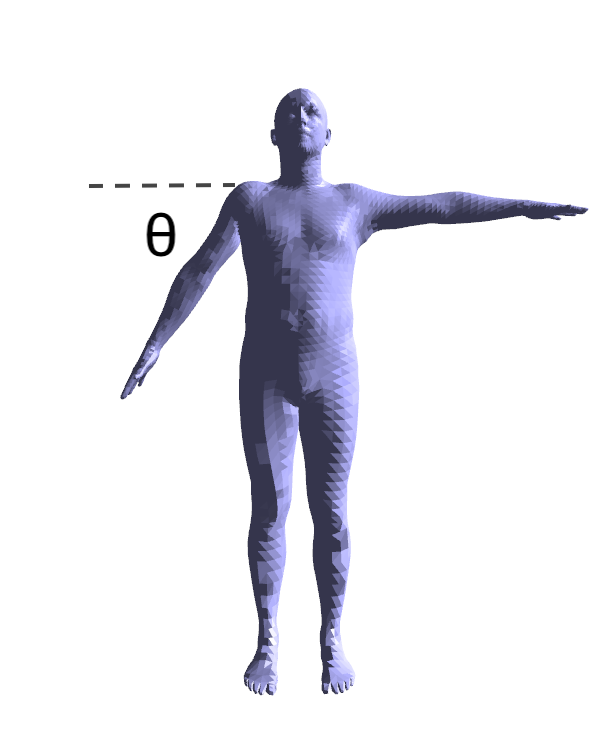}
        \vspace{-20pt}
        \caption{$M(\theta, \Ub)$}
		\label{fig:lbs_control}
    \end{subfigure}
    \begin{subfigure}[b]{0.15\textwidth}
        \centering
        \includegraphics[width=\textwidth]{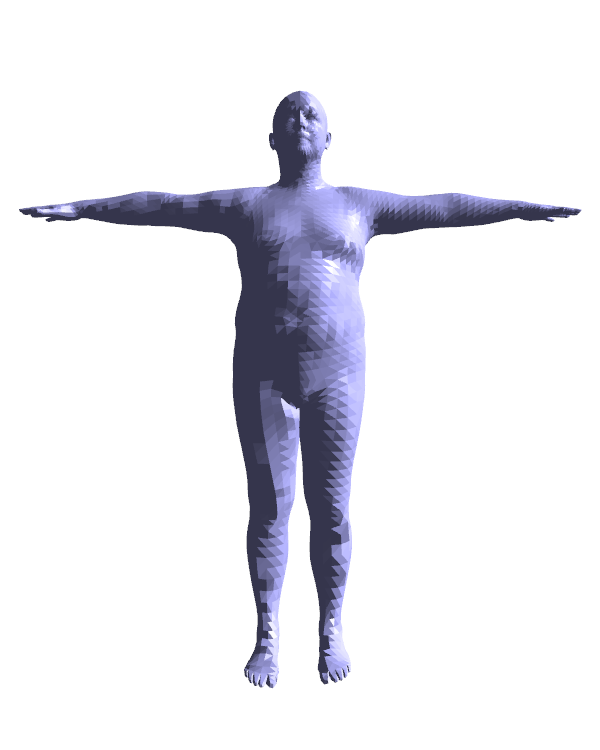}
        \vspace{-20pt}
        \caption{$\Ub^d$}
        \label{fig:deform}
    \end{subfigure}
    \vspace{-5pt}
    \caption{(a) Template mesh, (b) LBS deformation of the template using joint angles $\theta$, and (c) a deformed template.}
    \label{fig:lbs}
\end{figure}

\paragraph{Linear Blending Skinning} 
We start by briefly introducing LBS~\cite{magnenat1988joint}, which is the core building component of the proposed work. 
LBS models deformation of a mesh from a rest pose as a weighted sum of the
skeleton bone transformations applied to each vertex.
We follow the notation outlined in~\cite{loper2015smpl}, which is a strong influence on our model. An LBS model with $J$ joints can be defined as follows 
\begin{equation}
    \Vb = M(\Theta, \Ub),
    \label{eq:lbs}
\end{equation}
with $\Vb$ the vertices of the deformed shape after LBS. 
The LBS function $M$ takes two parameters, one is the vertices $\Ub=\{u_i\}_{i=1}^{m}$  of a base mesh (template), and the other are  
the relative joint rotation angles $\Theta\in \RR^{J\times 3}$ for each joint $j$ 
with respect to its parents. If $\Theta=\mathbf{0}$, then $M(\mathbf{0}, \Ub)=\Ub$. 
Two additional parameters, the skinning weights $w$ and the joint hierarchy $K$, are required by LBS. We will consider them fixed by the artist-defined rig. In particular,
$w\in\RR^{m\times J}$ defines the weights of each vertex contributing to joint $j$ and $\sum_j w_{i,j}=1$ for all $i$.
$K$ is the joint hierarchy. 
Each vertex $v_i \in \Vb$ can then be written as 
\[
    v_i = \left( \Ib_3, \mathbf{0}\right)\cdot \sum_{j=1}^J w_{i,j} \Tcal_j(\Theta, K) \begin{pmatrix} u_i \\ 1 \end{pmatrix}, 
\]
where $\Tcal_j(\Theta, K)\in \mbox{SE}(3)$ is a transformation matrix for each joint $j$, which
encodes the transformation from the rest pose to the posed mesh in world coordinate, constructed by traversing the hierarchy $K$ from the root to $j$.
Since each $v_i$ is constructed by a sequence of linear operations, the LBS $M(\Theta, \Ub)$ is differentiable respect to
$\Theta$ and $\Ub$.
A simple example constructed from the LBS component in SMPL~\cite{loper2015smpl} is shown in Figure~\ref{fig:lbs_temp}
and~\ref{fig:lbs_control}.

In this work, both the joint angles and the template mesh used in the LBS function are produced by deep networks from the input point cloud data,
\begin{equation}
    \Vb = M(f( \Xb ) , d( \Xb, \Ub )),
    \label{eq:lbs_full}
\end{equation}
where we identify a \emph{joint angle estimation} network $f$, and a \emph{template deformation} network $d$ which we describe below.

\vspace{-5pt}
\paragraph{Joint Angle (Pose) Estimation}
Given an LBS model defined in~\eqref{eq:lbs}, the goal is to regress joint angles based on input $\Xb$ via a function $f: \Xb\rightarrow \Theta$ 
such that $M(f(\Xb), \Ub)\approx \Xb$. 
We use a deep neural network, which takes set data (\eg point cloud) as input~\cite{qi2017pointnet, zaheer2017deep} to $f$, but we must also specify how to compare $\Xb$ and $\Vb$ from $M(\cdot)$. 
Losses that assume uniformly sampled surfaces (such as distribution matching~\cite{li2018point} or optimal transport) are less suitable, because reconstructed point clouds typically exhibit some amount of missing data and non-uniform sampling.

Instead, we adopt a Chamfer distance (CD)~\cite{yang2018foldingnet} defined as  $\Lcal_c(\Xb, \Vb) = $
\begin{equation}
    \frac{1}{n}\sum_{i=1}^n \| x_i - \Ncal_\Vb(x_i) \|^2 + \frac{1}{m}\sum_{j=1}^m \| v_j - \Ncal_\Xb(v_j) \|^2,
    \label{eq:chamfer}
\end{equation}
where $\Ncal_\Vb(x_i)=\arg\min_{v_j\in \Vb} \|x_i-v_j\|$ is the nearest neighbor of $x_i$ in $\Vb$.
This is also called Iterative Closest Point (ICP) in the registration
literature~\cite{besl1992method}. 
After finding nearest neighbors, 
we learn $f$ by back-propagating this point-wise loss through the differentiable LBS $\Vb=M(f(\Xb),\Ub)$. 
Also note that we only sample a subset of points for estimating~\eqref{eq:chamfer} under SGD training schemes.

In practice, we observe that it takes many iterations for PointNet~\cite{qi2017pointnet} or DeepSet~\cite{zaheer2017deep} architectures to improve if the target loss is 
CD instead of corresponded supervision. Similar behaviors were observed in~\cite{yang2018foldingnet, li2018point}, where the algorithms may take millions of 
iterations to converge.
To alleviate this problem, we utilize LBS to generate data based on a given $\Theta'$ for \emph{self-supervision} by optimizing
\[
    \min_f \Lcal_\Theta = \|f( M(\Theta', \Ub) ) - \Theta' \|^2.
\]
It is similar to the \emph{loop-back} loss~\cite{genova2018unsupervised} that ensures $f$ can correctly reinterpret the model's own
output from $M$.
Different from~\cite{genova2018unsupervised, kanazawa2018end}, we do not assume a prior pose
distribution is available. Our $\Theta'$ comes from two sources of randomness. One is uniform distributions
within the given joint angle ranges (specified by the artist-defined rig) and the second is we uniformly perturb
the inferred angles from input samples with a small uniform noise on the fly, which can gradually
adapt to the training data distribution when the estimation is improved as training progresses (see Section~\ref{sec:scd} and Figure~\ref{fig:mix}).
\begin{figure}[t]
    \centering
    \includegraphics[width=1.01\linewidth]{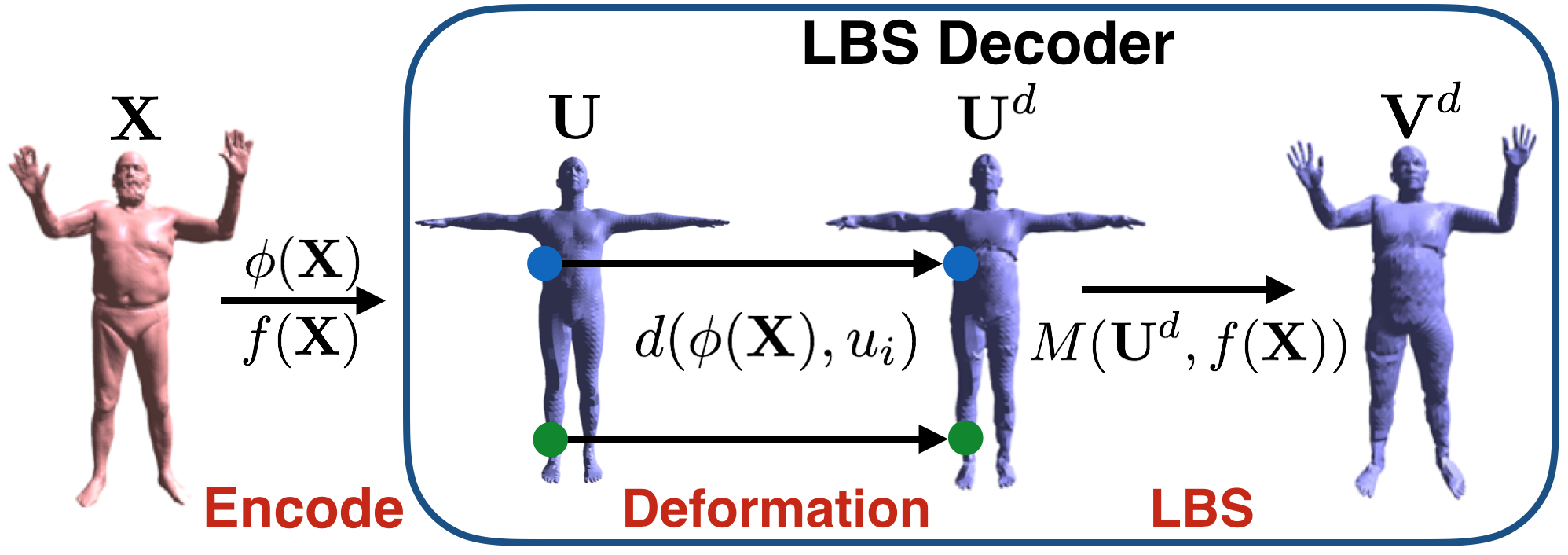}
	\caption{LBS-AE. Given a point cloud $\Xb$ of a input shape, we encode $\Xb$ into a latent code $\phi(\Xb)$ and the
	inferred joint angles $f(\Xb)$. The decoder contains a deformation network $d$ to deform the template $\Ub$ into
	$\Ub^d$, then uses a LBS to pose $\Ub^d$ into $\Vb^d$ as the reconstruction.}
	\label{fig:dec}
\end{figure}

\vspace{-8pt}
\paragraph{Template Deformation}
Although LBS can represent large pose deformations, due to limitations of LBS as well as differences between the artist
modeled mesh and the real data, there will be a large residual in the fitting. We refer to this residual as a
\emph{modality gap} between the model and reality, and alleviate this difference by using a neural network $d$ to produce 
the template mesh to be posed by LBS. 
The deformation network $d(\phi(\Xb), u_i)$ takes two sources as input, where $u_i$ is each vertex in 
the template mesh $\Ub$, and $\phi(\Xb)$ are features from an intermediate layer in $f$, which contains information about the state of $\Xb$.
This yields a deformed template $\Ub^d = \{d(\phi(\Xb), u_i)\}_{i=1}^m$.
One example is shown in Figure~\ref{fig:deform}.
After LBS, we denote the deformed and posed mesh as $\Vb^d = M(f(\Xb), \Ub^d)$, and denote by $\Vb=M(f(\Xb), \Ub)$ the posed original template.

If $d$ is high-capacity, $f(\Xb)$ can learn to generate all-zero joint angles for the LBS component (ignoring the input $\Xb$), and explain all deformations instead with $d$. That is, 
$M(f(\Xb), \Ub^d)=M(\mathbf{0}, \Ub^d)=\Ub^d \approx \Xb$, which reduces to the unsupervised version of~\cite{groueix20183d}. 
Instead of using explicit regularization to constrain $d$ (\eg $\|d(\phi(\Xb), \Ub^B)\|$), we propose a composition of two 
Chamfer distances as 
\begin{equation}
    \Lcal_{c^2,\lambda} =  \Lcal_c\left(\Xb, \Vb^d\right) + \lambda \Lcal_c\left(\Xb, \Vb\right). 
    \label{eq:c_chamfer}
\end{equation}
The second term in ~\eqref{eq:c_chamfer} enforces $f(\Xb)$ to learn correct joint angles even without template deformation.

Lastly, we follow~\cite{kanazawa2018learning, groueix20183d} and apply Laplacian regularization $\Lcal_{\text{\tt lap}} = \|L\Vb^d\|$ to encourage
smoothness of the deformed template, where $L$ is the discrete Laplace-Beltrami operator constructed from mesh $\Ub$ and its faces.

\vspace{-8pt}
\paragraph{LBS-based Autoencoder}
The proposed algorithm can be interpreted as an encoder-decoder scheme. The joint angle regressor is the encoder, which
compresses $\Xb$ into style codes $\phi(\Xb)$ and interpretable joint angles $f(\Xb)$. The decoder, different from
standard autoencoders, is constructed by combining a human designed LBS function and a style deformation network $d$ on the base template. We call the proposed algorithm LBS-AE as shown in Figure~\ref{fig:dec}.

\subsection{Structured Chamfer Distance}
\label{sec:scd}
To train an autoencoder, we have to define proper reconstruction errors for different data. 
In LBS-AE, the objective that provides information about input point clouds is only 
CD~\eqref{eq:chamfer}.
However, it is known that CD has many undesirable local optima, which hinders the algorithm from improving.

\begin{figure}
    \centering
    \begin{subfigure}[b]{0.10\textwidth}
        \centering
        \includegraphics[trim={50 10 10 10},clip,width=\textwidth]{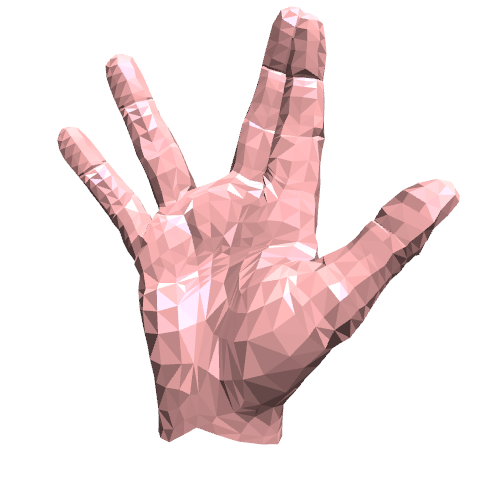}
        \vspace{-20pt}
        \caption{Input}
		\label{fig:local_t}
    \end{subfigure}
    \begin{subfigure}[b]{0.10\textwidth}
        \centering
        \includegraphics[trim={50 10 10 10},clip,width=\textwidth]{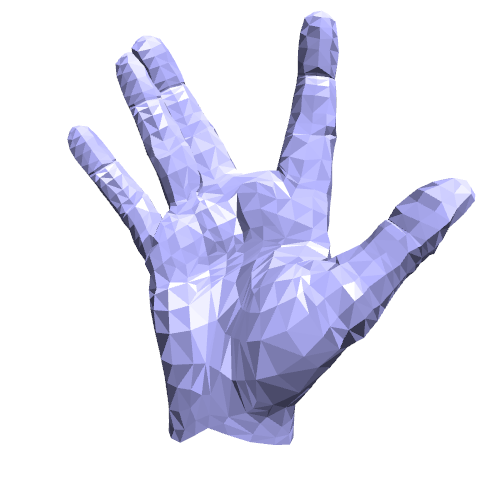}
        \vspace{-20pt}
        \caption{Estimate}
		\label{fig:local_e}
    \end{subfigure}
    \begin{subfigure}[b]{0.26\textwidth}
        \centering
        \includegraphics[width=\textwidth]{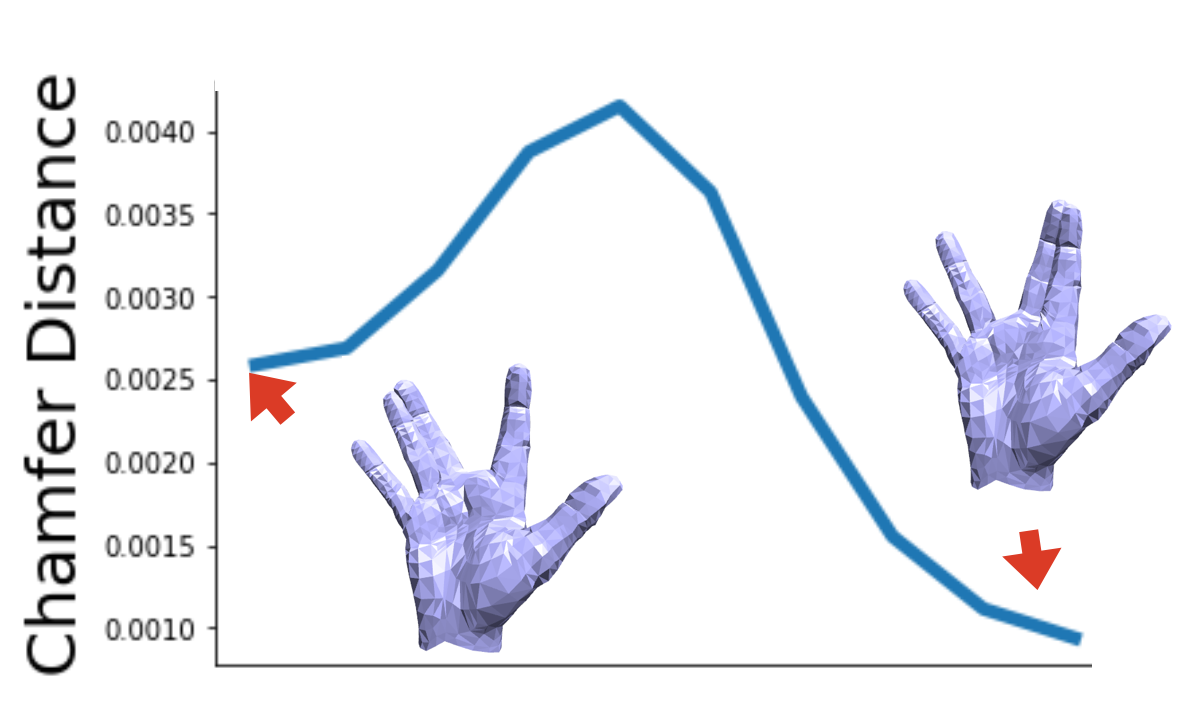}
        \label{fig:d_template}
        \vspace{-20pt}
        \caption{}
    \end{subfigure}
    \vspace{-5pt}
    \caption{When we try to move the 
    middle finger of the current estimate (b) toward the target (a), the Chamfer distance increases before decreasing, showing a local optimum that is difficult to overcome.}
    \label{fig:local}
    \vspace{-15pt}
\end{figure}

A local optimum example of CD is shown in Figure~\ref{fig:local}. 
To move the middle finger from the current estimate towards the index finger to fit the input, the Chamfer distance must increase before decreasing. 
This local optimum is caused by incorrect correspondences found by nearest neighbor search (the nearest neighbor of the middle finger of the current estimate is the
ring finger of the input). 

\paragraph{High-Level Correspondence}
Given a pair of sets $(\Vb, \Xb)$, for each $v \in \Vb$, we want to find its correspondence $\Ccal_\Xb(v)$ in $\Xb$. In
CD, we use the nearest neighbor $\Ncal_\Xb(v)$ to approximate $\Ccal_\Xb(v)$, which can be wrong, as shown in Figure~\ref{fig:local}. 
Instead of searching for nearest neighbors $\Ncal_\Xb(v)$ over
the entire set $X$, we propose to search within a subset $X' \subset X$, where $\Ccal_\Xb(v) \in X'$,  by eliminating 
irrelevant points in $\Xb$. 
Following this idea, we partition $\Xb$ into $k$ subsets, 
$\Xb^1 \dots \Xb^k$, where we use
$s(x; \Xb)\in \{1, \dots, k\}$ to denote which subset $x$ belongs to. 
A desirable partition should ensure $s(v; \Vb) =s(\Ccal_\Xb(v); \Xb)$; then, to find the nearest neighbor of $v$, we need only consider $\Xb^{s(v)}\subset \Xb$. 
We then define the \emph{Structured Chamfer Distance} (SCD) as $\Lcal_s(\Xb, \Vb)=$
\begin{equation}
    \frac{1}{n}\sum_{i=1}^n \| x_i - \Ncal_{\Vb^{s(x)}}(x_i) \|^2 +  \frac{1}{m}\sum_{j=1}^m \| v_j - \Ncal_{\Xb^{s(v)}}(v_j) \|^2,
    \label{eq:s_chamfer}
\end{equation}
where we ease the notation of $s(x, \Xb)$ and $s(v, \Vb)$ to be $s(x)$ and $s(v)$. 
Compared with CD, which finds nearest neighbors from \emph{all to all}, SCD uses \emph{region to region} based on the
high-level correspondence by  leveraging the structure of data.
Similar to~\eqref{eq:c_chamfer}, we define
\begin{equation}
    \Lcal_{s^2,\lambda} =  \Lcal_s\left(\Xb, \Vb^d\right) + \lambda \Lcal_s\left(\Xb, \Vb\right). 
    \label{eq:c_s_chamfer}
\end{equation}

\setlength{\columnsep}{5pt}
\begin{wrapfigure}[7]{r}{0.45\linewidth}
    \centering
    \vspace{-10pt}
    \includegraphics[clip,width=1\linewidth]{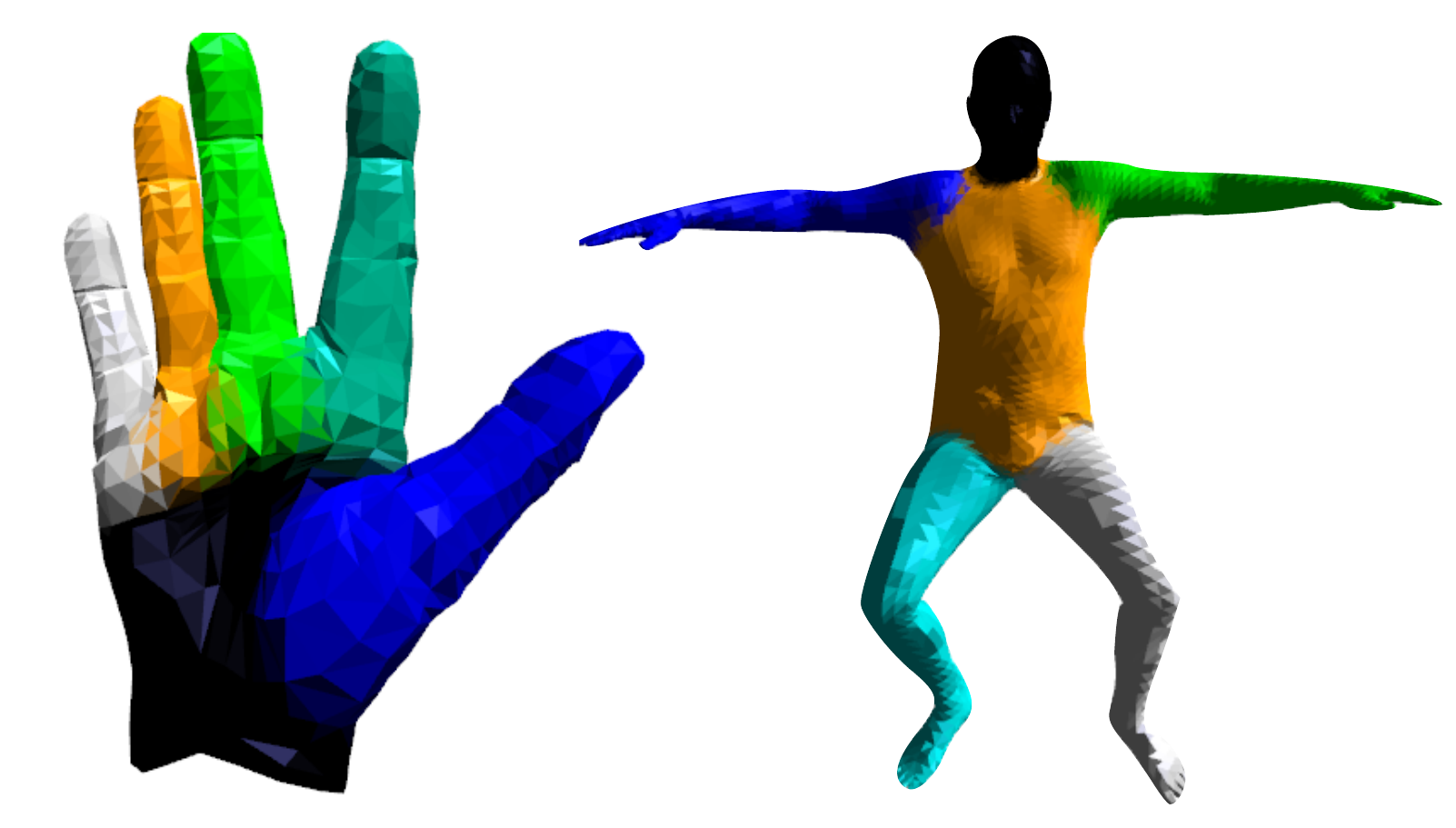}
    \vspace{-18pt}
    \caption{Joint Partitions.}
    \label{fig:partition}
\end{wrapfigure}
In this paper, we partition the vertices based on the LBS skinning weights at a chosen granularity. Examples of hand and body data are shown
in Figure~\ref{fig:partition}, which use the structure and our prior knowledge of the human body. 
These
satisfy the property that the true correspondence is within the 
same partition. 
With the proposed SCD, we can improve the local optimum in Figure~\ref{fig:local}. 

\begin{figure}[t]
    \centering
    \includegraphics[width=1.0\linewidth]{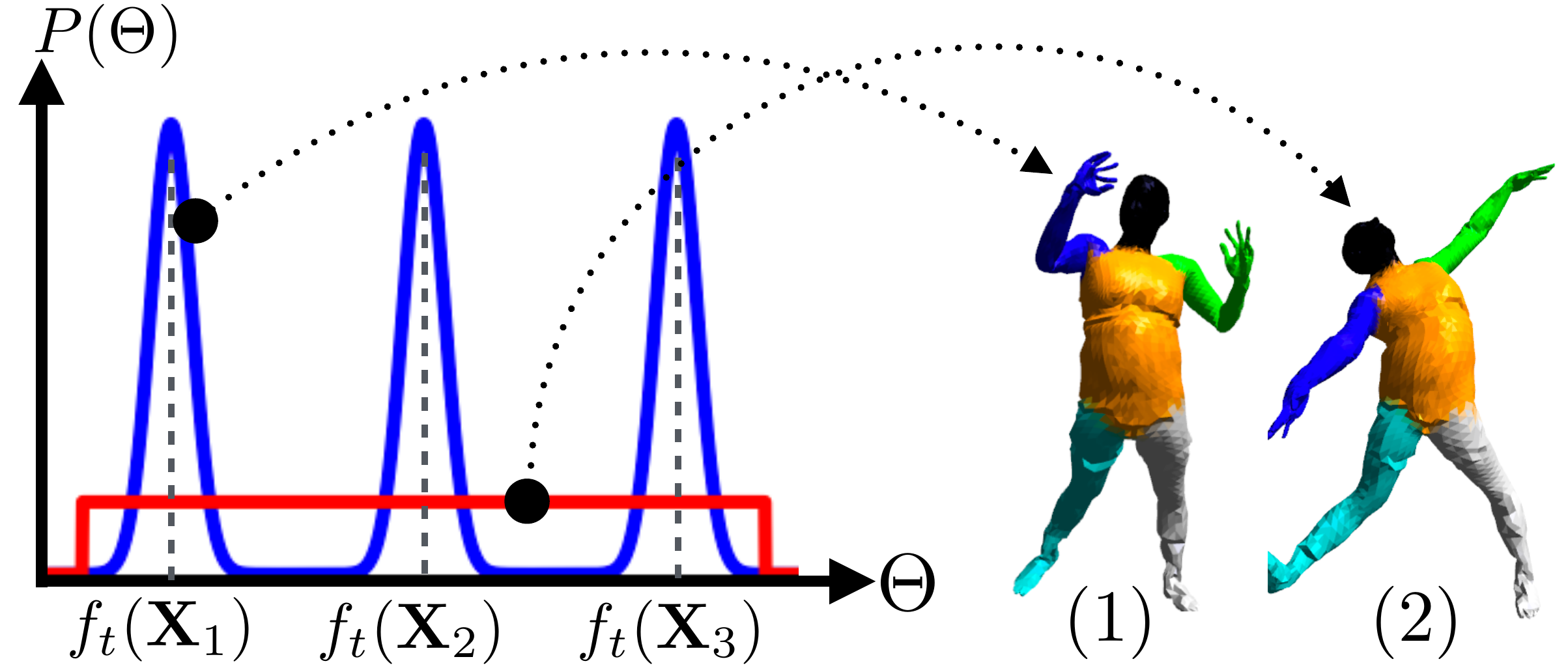}
	\caption{The mixture distribution of self-supervision data from the LBS at iteration $t$. 
	We sample from (1) the perturbed distribution centered at the $f_t(\Xb_i)$ and (2) a uniform distribution.}
	\label{fig:mix}
\end{figure}

\paragraph{Segmentation Inference}
For the deformed mesh $\Vb$, we can easily infer the partition $s(v;\Vb)$, because the mapping between vertices and joints is defined by the LBS skinning weights $w$.
We directly use $\argmax_j w_{i,j}$ as labels.
Without additional labeling or keypoint information, the difficulty is to infer $s(x;\Xb)$ for $x \in \Xb$, which is a point cloud segmentation task~\cite{qi2017pointnet}. 
However, without labels for $\Xb$, we are not able to train a segmentation model on $\Xb$ directly. Instead,
similar to the self-supervision technique used for 
training the joint angle regressor, we propose to train a segmentation network $s$ with the data $(\Vb^d, \Yb)$ generated by LBS, where 
$\Yb$ are the labels for $w$ defined in LBS and $\Vb=M(\Theta, \Ub^d)$.  Note that $\Theta$ follows the same distribution as before, which contains uniform sampling for exploration 
and perturbation of the inferred angles $f(\Xb)$, as shown in Figure~\ref{fig:mix}.
Instead of using the base template $\Ub$ only, we use the inferred deformed template $\Ub^d$ to adapt to the real data modality, which improves performance (see Section \ref{sec:seg_exp}).

The final objective for training the shape deformation pipeline including $f(\cdot)$ and $d(\cdot)$ is\footnote{We use $\lambda=0.5$, $\lambda_{\text{\tt lap}}=0.005$, $\lambda_\theta=0.5$ in all experiments.}
\begin{equation}
    \Lcal = \Lcal_{c^2, 0.5} + \lambda_s \Lcal_{s^2, 0.5} + \lambda_{\text{\tt lap}} \Lcal_{\text{\tt lap}} + \lambda_{\theta} \Lcal_{\Theta},
    \label{eq:all}
\end{equation}
and we use standard cross-entropy for training $s$. 
In practice, since $s$ is noisy during the first iterations, we pretrain it for $50K$ 
iterations with poses from uniform distributions over joint angles. 
Note that, for pretraining, we can only use the base template $\Ub$ to synthesize data. 
After that, we then learn everything jointly by updating each network alternatively. 
The final algorithm, LBS-AE with SCD as reconstruction loss, is shown in Algorithm~\ref{algo:lbsae}.

\begin{algorithm}[t]
\caption{LBS-AE with SCD}
\label{algo:lbsae}
\begin{algorithmic} 
    \STATE\textbf{Inputs:} \textbullet~Point Clouds: $\{\Xb\}$ \\
    \hspace{33pt}\textbullet~LBS: $M(;w, K, \Ub)$ and angle ranges $(R_l, R_u)$
    \vspace{-10pt}
    \STATE Pretrain $s$ on uniformly sampled poses from LBS
    \STATE\textbf{while} $f$ and $d$ have not converged: 
        \STATE\hspace{8pt}1. Sample minibatch $\{\Xb_i\}_{i=1}^B$, $\{\Xb_i'\}_{i=1}^B$
        \STATE\hspace{8pt}2. $\Theta'= \{ f(\Xb_i)+\epsilon_i \}_{i=1}^B \cup \Theta_r\sim \mbox{Unif}(R_l,R_u)$
        \STATE\hspace{8pt}3. Generate $(\Vb^d, \Yb)$ based on $\Theta'$ to update $s$
        \STATE\hspace{8pt}4. Infer segmentation labels $\{s(\Xb_i')\}_{i=1}^B$ 
        \STATE\hspace{8pt}5. Update $f$ and $d$ based on (1)-(3) (Eq.~\eqref{eq:all}) 
\end{algorithmic}
\end{algorithm}

\section{Related Works}

\paragraph{LBS Extensions}
Various extensions have been proposed to fix some of the shortcomings of LBS~\cite{lewis2000pose, sloan2001shape, wang2002multi, kavan2005spherical, rhee2006real, joshi2007harmonic, kavan2008geometric, le2012smooth, zuffi2015stitched, loper2015smpl, bailey2018fast}, 
where we only name afew here.  
The proposed template deformation follows the idea of~\cite{kurihara2004modeling, rhee2006real, zuffi2015stitched, loper2015smpl} to model the modalities and corrections 
of LBS on the base template rest pose. \cite{zuffi2015stitched,loper2015smpl} use PCA-like algorithms to model modalities via a weighted sum of learned shape basis. Instead,  our approach is similar to~\cite{bailey2018fast} by learning modalities via a deformation network.
The main difference between LBS-AE and~\cite{zuffi2015stitched, loper2015smpl, bailey2018fast} is we do not rely on correspondence information to learn the template deformation $d$ a priori. 
We simultaneously learn $d$ and infer pose parameters without external labeling.
\vspace{-10pt}
\paragraph{Deep Learning for 3D Data}
Many deep learning techniques have been developed for different types of 3D information, such as 3D voxels~\cite{girdhar2016learning, wu20153d,wu2016learning}, 
geometry images~\cite{sinha2016deep, sinha2017surfnet}, meshes~\cite{bronstein2017geometric}, depth maps~\cite{wang2016action} and 
point clouds~\cite{qi2017pointnet, qi2017pointnet++, zaheer2017deep}. 
Autoencoders for point clouds are explored by~\cite{yang2018foldingnet, groueix2018atlasnet, li2018point, achlioptas2018learning}.

\vspace{-10pt}
\paragraph{Model Fitting with Different Knowledge}
Different works have studied to registration via fitting a mesh model by leveraging different levels of information about the data. 
\cite{kanazawa2018end} use SMPL~\cite{loper2015smpl} to reconstruct meshes from images by using key points and prior knowledge of distributions of pose parameters.
\cite{kanazawa2018learning} explore using a template instead of a controllable model to reconstruct the mesh with key points.
\cite{bogo2016keep, joo2018total} also adopt pretrained key point detectors from other sources of data as supervision.
Simultaneous training to improve model fitting and key point detection are explored by~\cite{lassner2017unite, mehta2017vnect}. 
The main difference from the proposed joint training in LBS-AE is we do not rely on an additional source of real-world data to pretrain networks, as needed to train these key point detectors. 
\cite{wei2016dense} share a similar idea of using segmentation for nearest neighbor search, but they trained the segmentation from labeled examples.
\cite{genova2018unsupervised} propose to control morphable models instead of rig models for modeling faces. They also utilize prior knowledge of the 
3DMM parameter distributions for real faces.  
We note that most of the works discussed above aim to recover 3D models from images.
\cite{groueix20183d} is the most related work to the proposed LBS-AE, but doesn't use LBS-based deformation. They use a base template and learn the full deformation process with a

neural network trained by correspondences provided a priori or from nearest neighbor search. More comparison between~\cite{groueix20183d} and LBS-AE will be studied in Section~\ref{sec:exp}.
Lastly, learning body segmentation via SMPL is studied by~\cite{varol2017learning}, but with a focus on learning a segmentation using SMPL with parameters inferred from real-world 
data to synthesize training examples.

\vspace{-5pt}
\paragraph{Loss Function with Auxiliary Neural Networks}
Using auxiliary neural networks to define objectives for training targeted models is also broadly studied in 
GAN literature (\eg~\cite{goodfellow2014generative, mao2016least, nowozin2016f, arjovsky2017wasserstein, li2017mmd, mroueh2017fisher, gulrajani2017improved}). 
\cite{li2018point} use a GAN loss for matching input and reconstructed point clouds. By leveraging prior knowledge, the auxiliary network adopted by LBS-AE is an interpretable segmentation network which can be 
trained without adversarial training. 

\section{Experiment}
\label{sec:exp}
\begin{figure}
    \centering
    \includegraphics[trim={120 10 90 0},clip,width=0.185\linewidth]{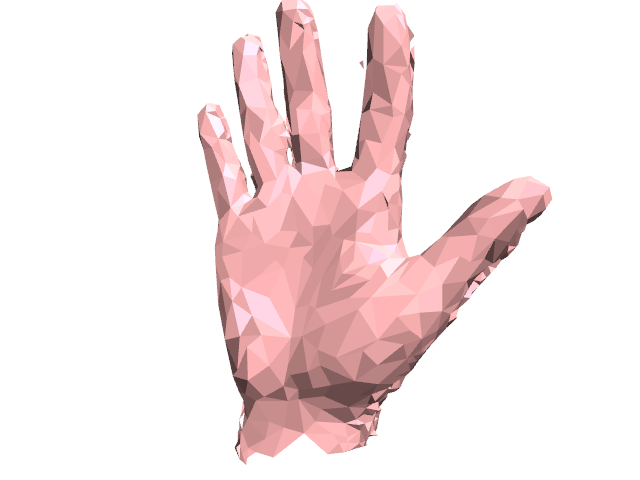}
    \includegraphics[trim={180 100 180 100},clip,width=0.185\linewidth]{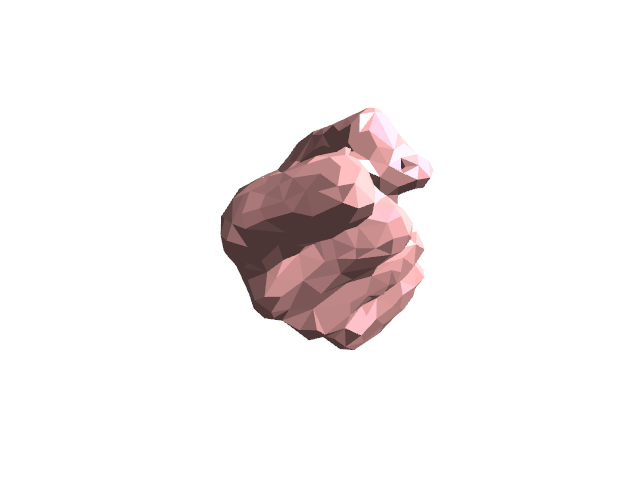}
    \includegraphics[trim={200 100 60 100},clip,width=0.185\linewidth]{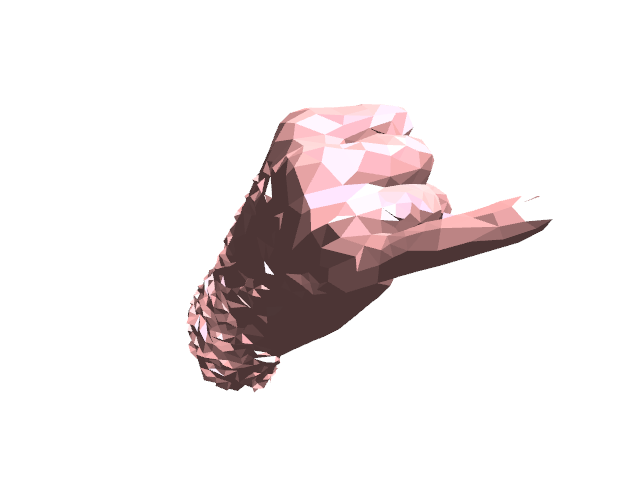}
    \includegraphics[trim={200 100 60 0},clip,width=0.185\linewidth]{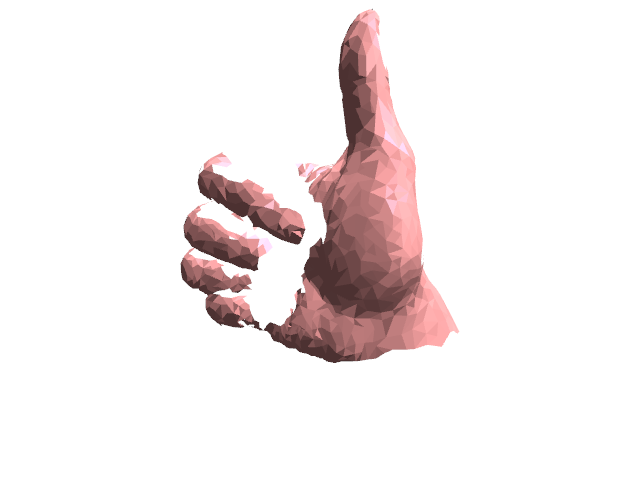}
    \includegraphics[trim={180 100 100 20},clip,width=0.185\linewidth]{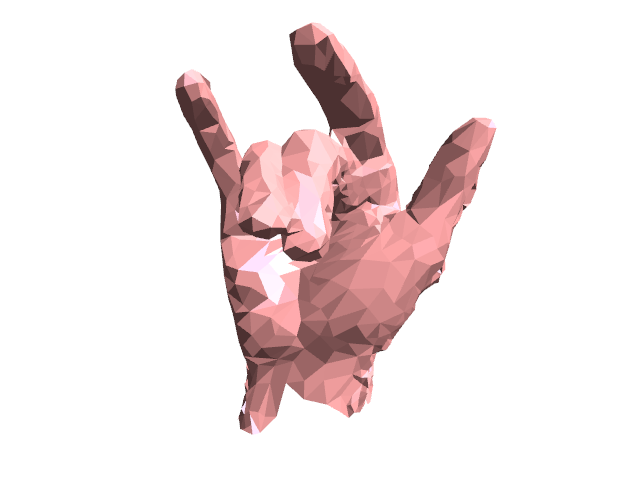}
    \vspace{-8pt}
    \caption{Examples of the captured hands.}
    \vspace{-12pt}
    \label{fig:hand}
\end{figure}

\paragraph{Datasets}
We consider hand and body data.
For body data, we test on FAUST benchmark~\cite{bogo2014faust}, which captures real human body with correspondence labeling.
For hand data, we use a multi-view capture system to captured $1,524$ poses from three people, which have missing area and
different densities of points across areas.
The examples of reconstructed meshes are shown in Figure~\ref{fig:hand}.
For numerical evaluation, 
in addition to FAUST,
we also consider synthetic data since we do not have labeling information on the hand data (\eg key points, poses, correspondence). 
To generate synthetic hands, we first estimate  pose parameters of the captured data under LBS. To model the modality gap, we prepare different base templates with various thickness and length of palms and fingers. We then generate data with LBS based on those templates and the inferred pose parameters.
We also generate synthetic human body shapes using SMPL~\cite{bogo2016keep}. 
We sample $20,000$ parameter configurations estimated by SURREAL~\cite{varol2017learning} and $3,000$ samples of bent shapes from~\cite{groueix20183d}.
For both synthetic hand and body data, 
the scale of each shape is in $[-1, 1]^3$ and we generate $2300$ and $300$ examples as holdout testing sets.

\vspace{-10pt}
\paragraph{Architectures} 
The architecture of $f$ follows~\cite{li2018point} to use DeepSet~\cite{zaheer2017deep}, which shows competitive performance with PointNet~\cite{qi2017pointnet} 
with half the number of parameters. The output is set to be $J\times 3$ dimensions, where $J$ is the number of joints. We use the previous layer's activations as $\phi(\Xb)$ for $d$. 
We use a three layer MLP to model $d$, where the input is the concatenation of $v$, $f(\Xb)$ and $\phi(\Xb)$, and the hidden layer sizes are $256$ and $128$.
For segmentation network $s$, we use~\cite{qi2017pointnet} because of better performance. 
For hand data, we use an artist-created LBS, while we use the LBS part from SMPL~\cite{loper2015smpl} for body data.

\subsection{Study on Segmentation Learning} 
\label{sec:seg_exp}
\begin{figure}
    \centering
    \begin{subfigure}[b]{0.152\textwidth}
        \centering
        \includegraphics[width=\textwidth]{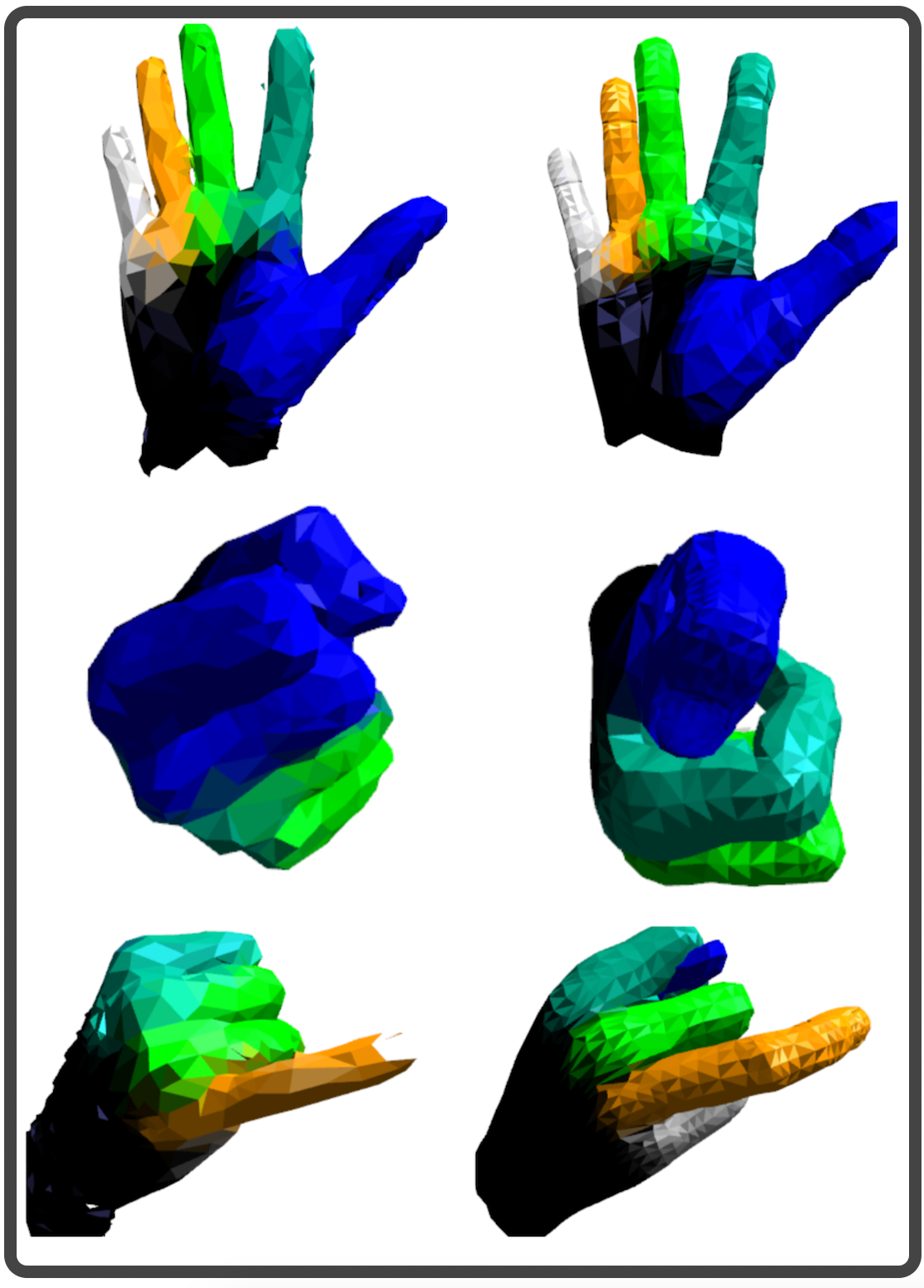}
        \caption{LBS-AE$_{\text{CD}}$}
        \label{fig:lbsae_cd}
    \end{subfigure}
    \begin{subfigure}[b]{0.152\textwidth}
        \centering
        \includegraphics[width=\textwidth]{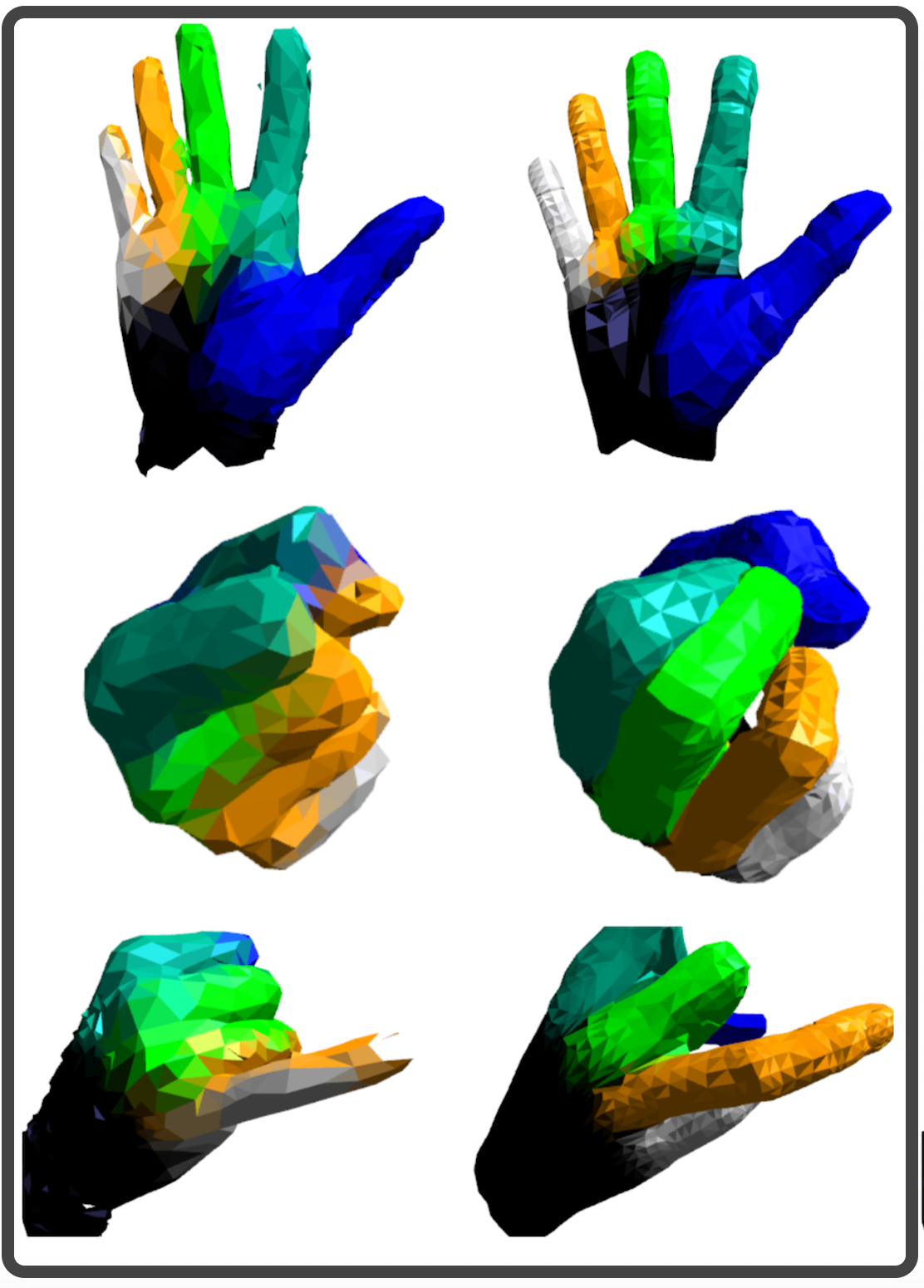}
        \caption{Modality Gap}
        \label{fig:modality}
    \end{subfigure}
    \begin{subfigure}[b]{0.15\textwidth}
        \centering
        \includegraphics[width=\textwidth]{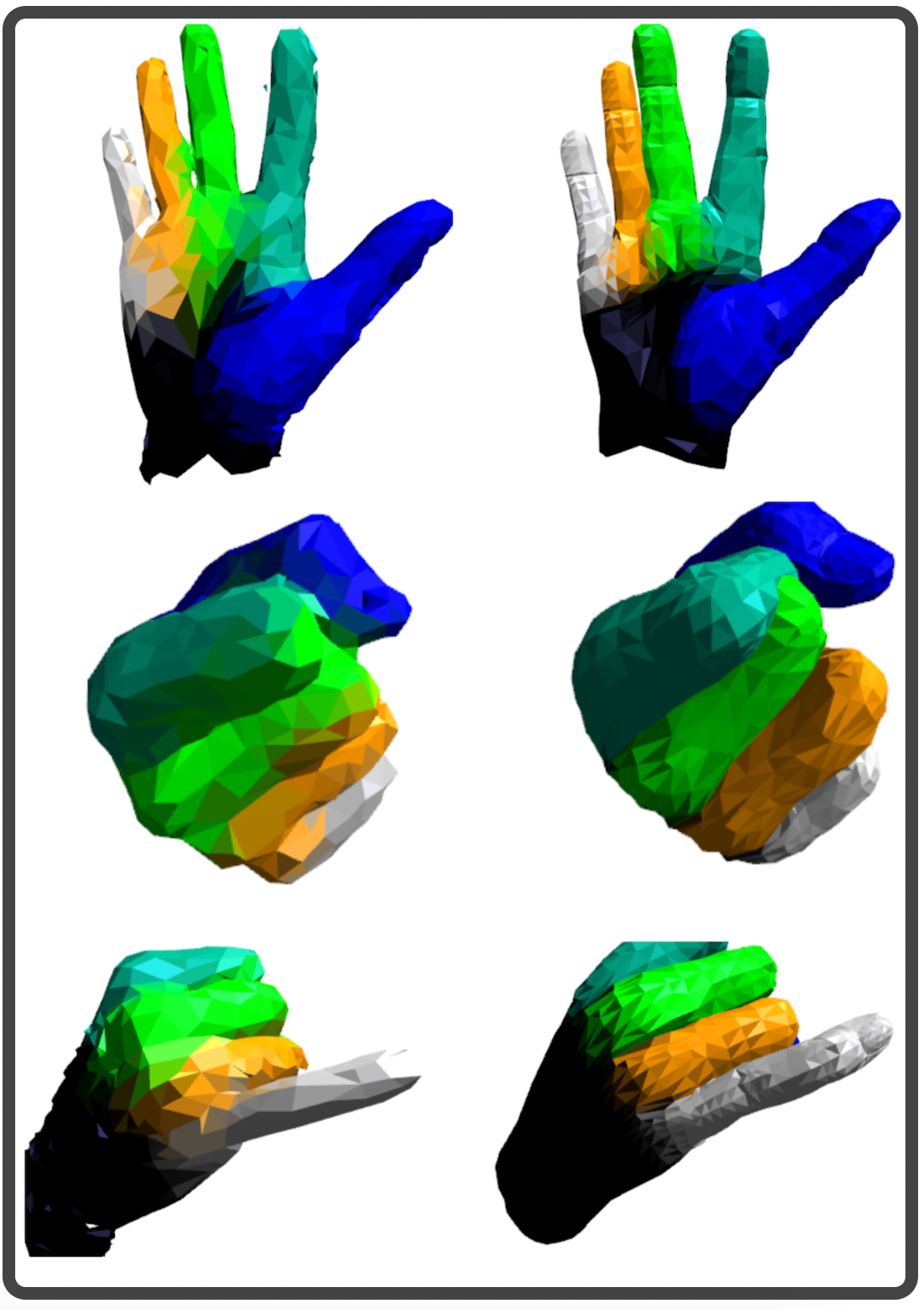}
        \caption{LBS-AE}
        \label{fig:hand_lbs}
    \end{subfigure}
    \vspace{-8pt}
    \caption{Ablation study of the proposed LBS-AE. For each block, the left column is the inferred segmentations of
	input shapes while the right column is the reconstruction.}
    \vspace{-12pt}
    \label{fig:loo}
\end{figure}

One goal of the proposed LBS-AE is to leverage geometry structures of the shape, by learning segmentation jointly to 
improve correspondence finding via nearest neighbor searching when measuring the 
difference between two shapes. Different from previous works (\eg \cite{wei2016dense}), 
we do not rely on any human labels. 
We study how the segmentation learning with self-supervision interacts with the model fitting to data. 
We train different variants of LBS-AE to fit the captured hands data.
The first is learning LBS-AE with CD only (LBS-AE$_{\text{CD}}$). The objective is \eqref{eq:all} without $\Lcal_{s^2, 0.5}$.
We then train the segmentation network $s$ for SCD with hand poses sampled from uniform distributions based on $\Ub$ 
instead of $\Ub^d$.
Note that there is no interaction between learning $s$ and the other networks $f$ and $d$. 
The segmentation and reconstructed results are shown in Figure~\ref{fig:lbsae_cd}.
We observe that the segmentation network trained on randomly sampled poses from a uniform distribution can only segment easy poses correctly and 
fail on  challenging cases, such as feast poses, because of the difference between true pose distributions and
the uniform distribution used as well as the modality gaps between 
real hands and synthetic hands from LBS. 
On the other hand, LBS-AE$_{\text{CD}}$ is stuck at different local optimums. For example, it recovers to stretch the ring
finger instead of the little finger for the third pose.

Secondly, we study the importance of adapting to different modalities.
In Figure~\ref{fig:modality}, we train segmentation and LBS fitting jointly with SCD.
However, when we augment the data for training segmentation, 
we only adapt to pose distributions via $f(\Xb)$, instead of using the deformed $\Ub^d$.
Therefore, the training data for $s$ for this case has a modality gap between it and the true data. 
Compared with Figure~\ref{fig:lbsae_cd}, the joint training  benefits the performance, for example, on the feast pose.  
It suggests how good segmentation learning benefits reconstruction.  
Nevertheless, it still fails on the third pose. 
By training LBS-AE and the segmentation jointly with inferred modalities and poses, we could fit the poses better as shown in Figure~\ref{fig:hand_lbs}. This difference demonstrates the importance 
of training segmentation adapting to the pose distributions and different modalities. 

\begin{figure}
    \centering
    \begin{subfigure}[b]{0.23\textwidth}
        \centering
        \includegraphics[width=\textwidth]{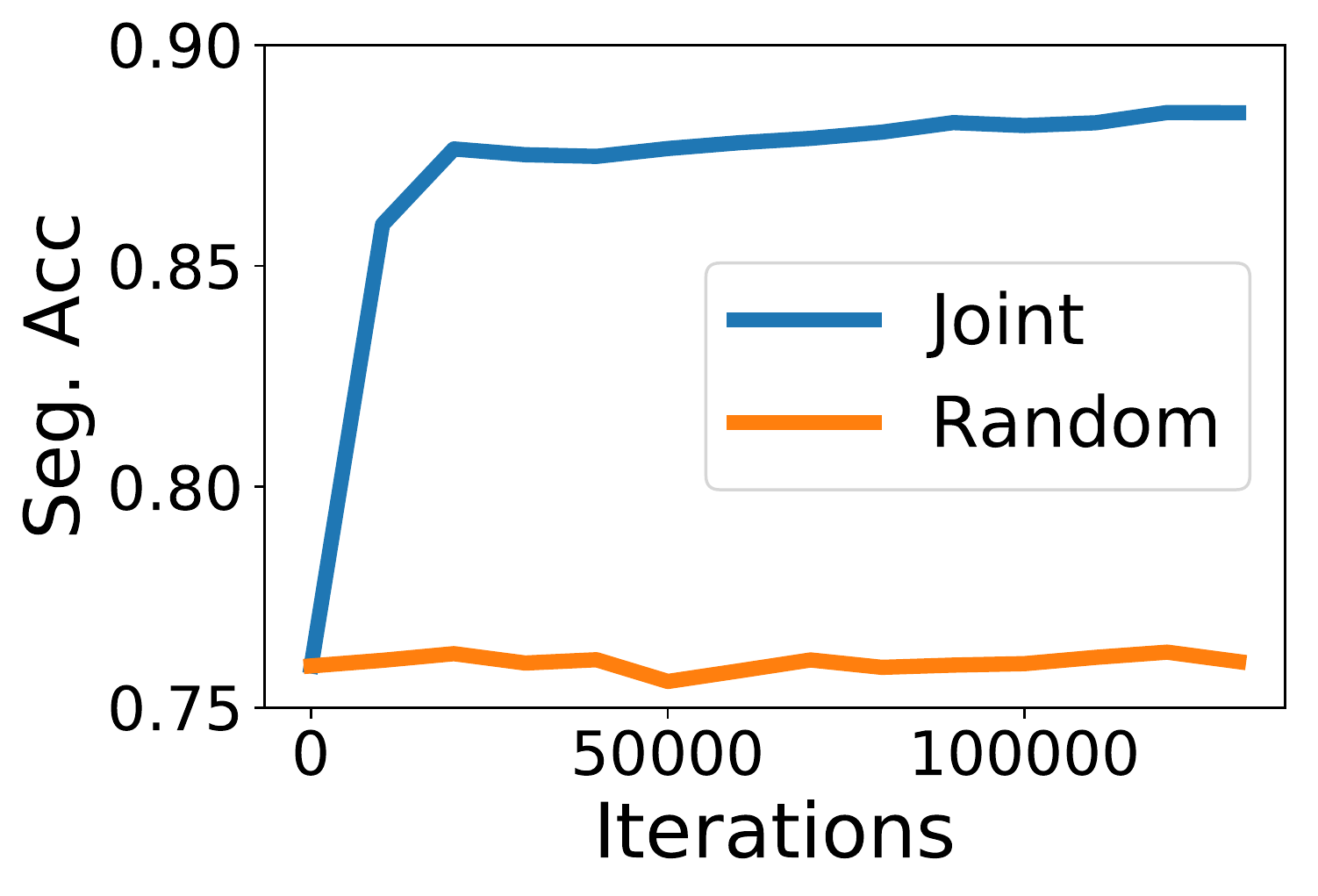}
        \caption{Synthetic Hands}
        \label{fig:hand_acc}
    \end{subfigure}
    \begin{subfigure}[b]{0.23\textwidth}
        \centering
        \includegraphics[width=\textwidth]{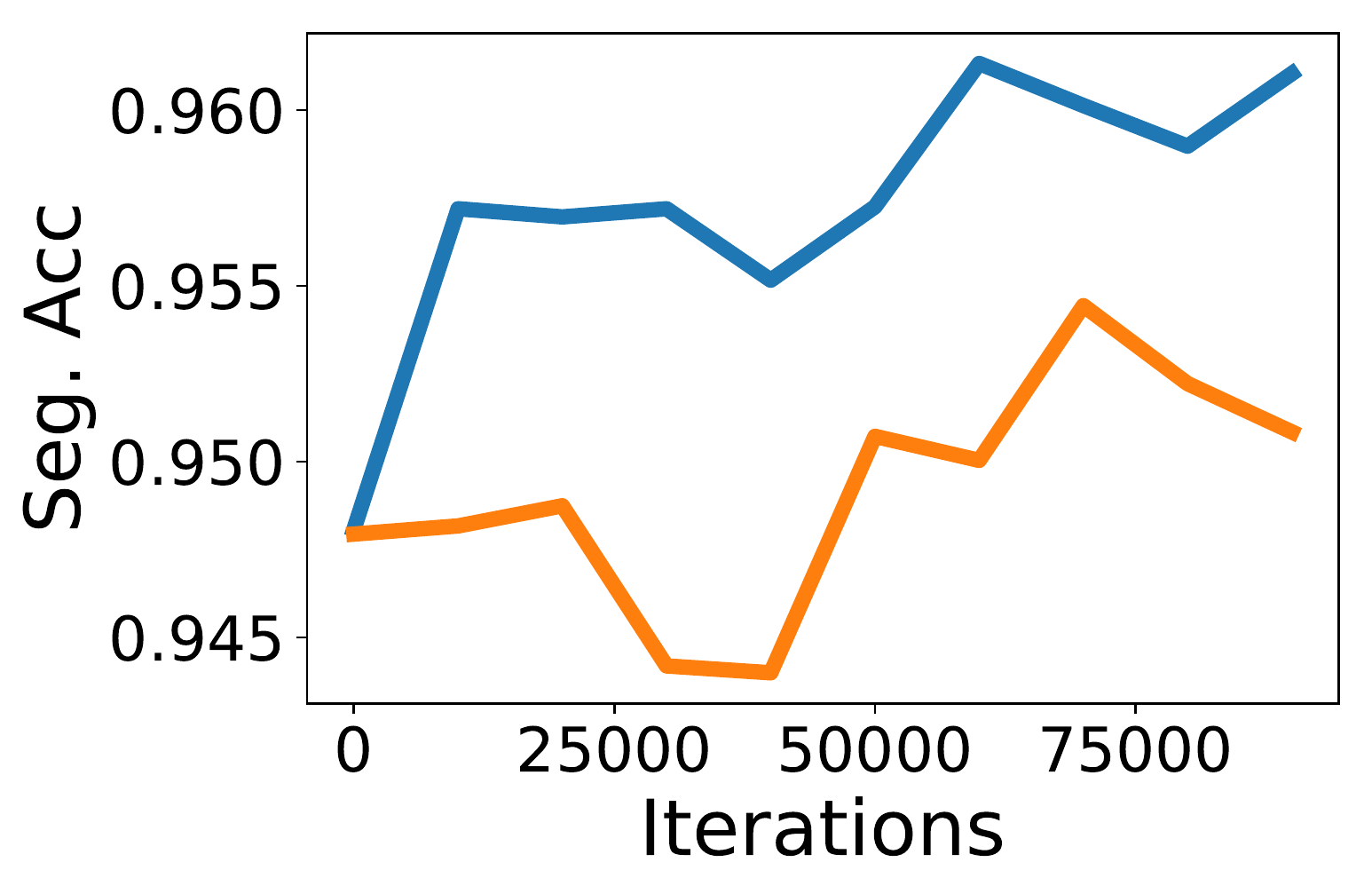}
        \caption{SMPL}
        \label{fig:body_acc}
    \end{subfigure}
    \vspace{-8pt}
    \caption{Segmentation accuracy on holdout testing sets.}
    \vspace{-10pt}
    \label{fig:seg_acc}
\end{figure}

\vspace{-10pt}
\paragraph{Numerical Results}
We also quantitatively investigate the learned segmentation when ground truth is available. We train $s$ with (1) randomly sampled shapes from uniform distributions over joint angle 
ranges ({\tt Random})
and (2) the proposed joint training ({\tt Joint}). 
We use pretraining as initialization as describing in Section~\ref{sec:scd}. We then 
train these two algorithms on the synthetic hand and body data and evaluate segmentation accuracy on the testing sets.
The results are shown in Figure~\ref{fig:seg_acc}.
{\tt Random} is exactly the same as pretraining. After pretraining, {\tt Random} is almost converged. 
On the other hand, {\tt Joint} improves the segmentation accuracy in both cases by gradually 
adapting to the true pose distribution when the joint angle regressor$f$ is improved. 
It justifies the effectiveness of the proposed joint training where we can infer the segmentation 
in a self-supervised manner.
For hand data, as we show in Figure~\ref{fig:loo}, there are many \emph{touching-skin} poses where fingers are touched to each other.
For those poses, there are strong correlations between joints in each pose, 
which are hard to be sampled by a simple uniform distribution and results in a performance gap in Figure~\ref{fig:hand_acc}. 
For body data, many poses from SURREAL are with separate limbs, which {\tt Random} can generalize surprisingly well. Although it seems {\tt Joint} only leads to incremental improvement over 
{\tt Random}, we argue this gap is substantial, especially for resolving challenging touching-skin cases as we will show
in Section~\ref{sec:quant}.

\subsection{Qualitative Study}

We compare the proposed algorithm with the unsupervised learning variant of~\cite{groueix20183d}, which learns the deformation by entirely relying on neural networks. 
Their objective is similar to \eqref{eq:all}, but using CD and Laplacian regularization only. For  fair comparison, we also generate synthetic data on the fly 
with randomly sampled poses and correspondence for~\cite{groueix20183d}, which boosts its performance. 
We also compare with the simplified version of the proposed algorithm by using CD instead of SCD, which is denoted as LBS-AE$_{\text{CD}}$ as above.

\begin{figure}
    \centering
    \begin{subfigure}[b]{0.091\textwidth}
        \centering
        \includegraphics[trim={200 100 100 0},clip,width=\textwidth]{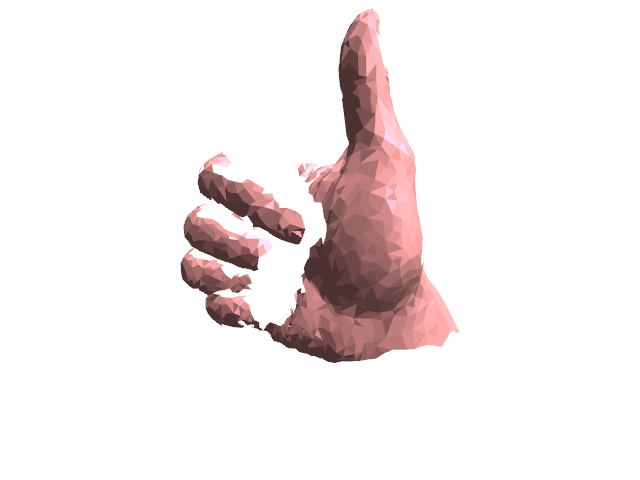}
    \end{subfigure}
    \begin{subfigure}[b]{0.091\textwidth}
        \centering
        \includegraphics[trim={200 100 100 0},clip,width=\textwidth]{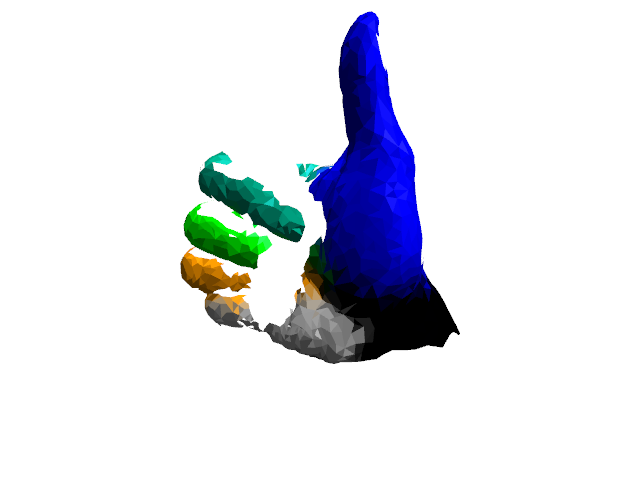}
    \end{subfigure}
    \begin{subfigure}[b]{0.091\textwidth}
        \centering
        \includegraphics[trim={200 100 100 0},clip,width=\textwidth]{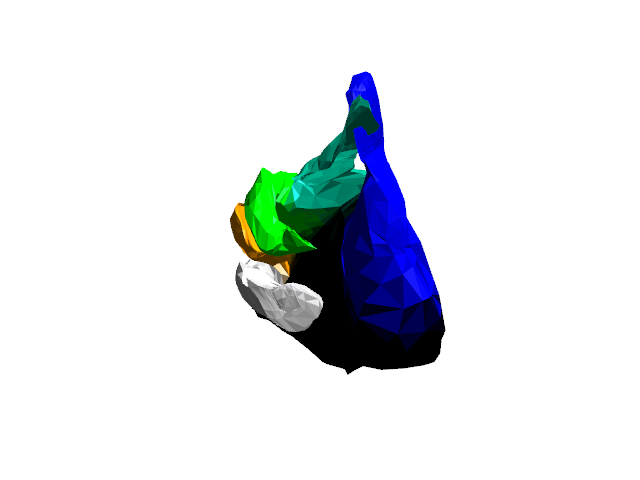}
    \end{subfigure}
    \begin{subfigure}[b]{0.091\textwidth}
        \centering
        \includegraphics[trim={200 100 100 0},clip,width=\textwidth]{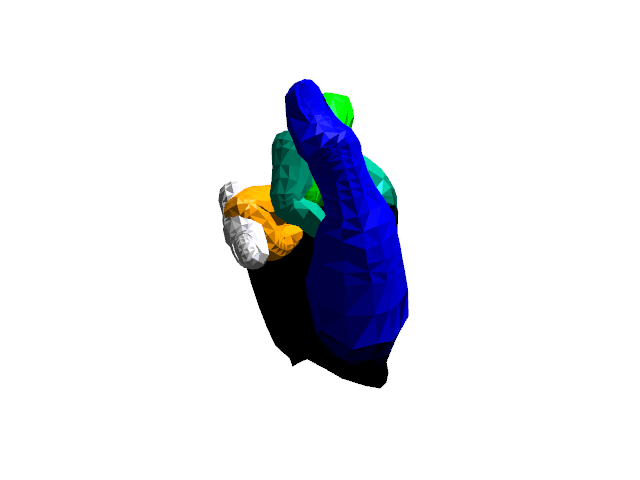}
    \end{subfigure}
    \begin{subfigure}[b]{0.091\textwidth}
        \centering
        \includegraphics[trim={200 100 100 0},clip,width=\textwidth]{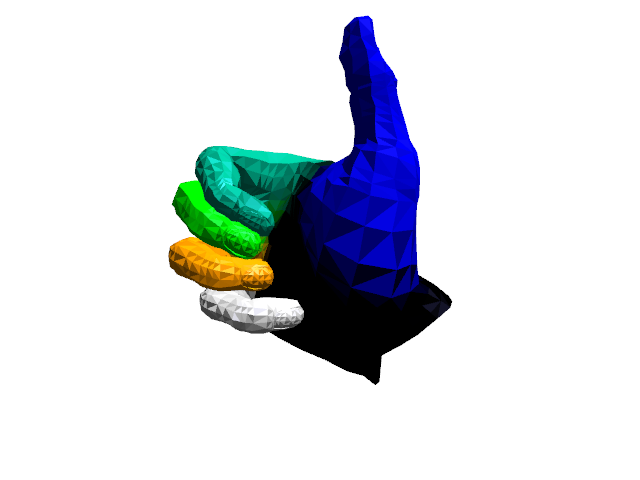}
    \end{subfigure}
    \begin{subfigure}[b]{0.091\textwidth}
        \centering
        \includegraphics[trim={180 50 100 0},clip,width=\textwidth]{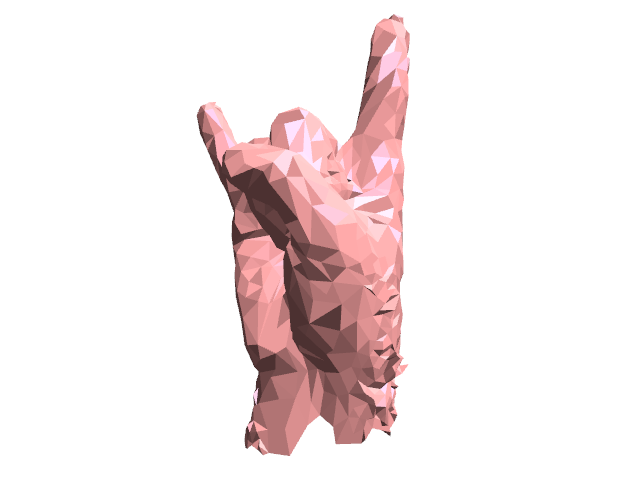}
    \end{subfigure}
    \begin{subfigure}[b]{0.091\textwidth}
        \centering
        \includegraphics[trim={180 50 100 0},clip,width=\textwidth]{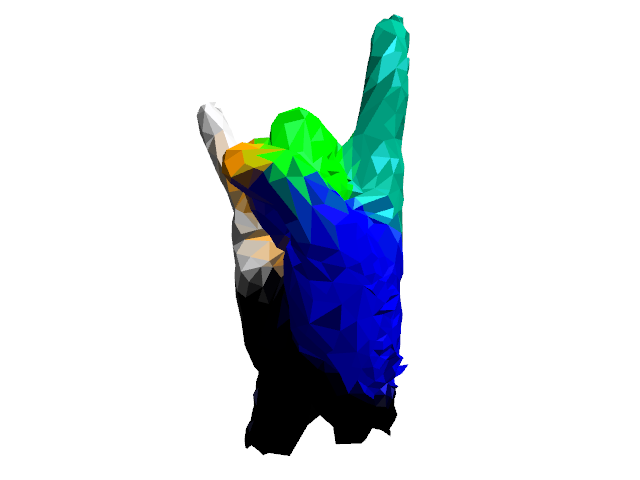}
    \end{subfigure}
    \begin{subfigure}[b]{0.091\textwidth}
        \centering
        \includegraphics[trim={180 50 100 0},clip,width=\textwidth]{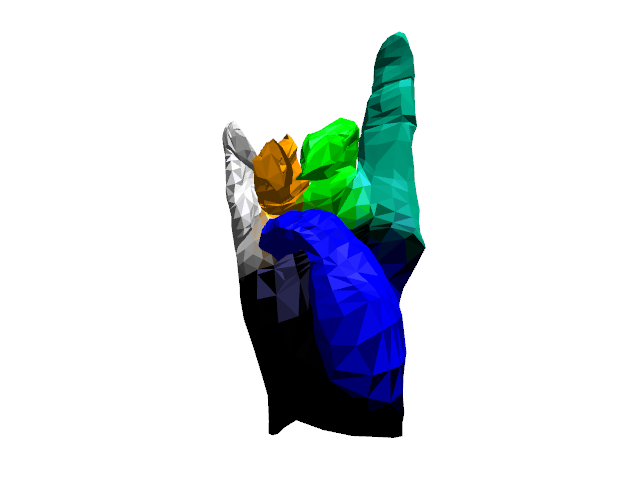}
    \end{subfigure}
    \begin{subfigure}[b]{0.091\textwidth}
        \centering
        \includegraphics[trim={180 50 100 0},clip,width=\textwidth]{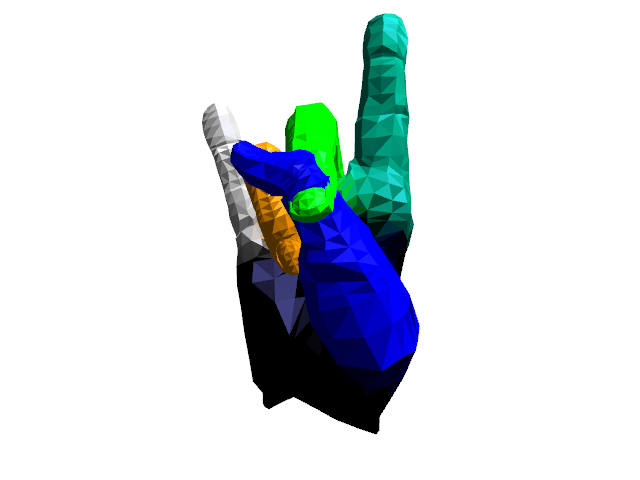}
    \end{subfigure}
    \begin{subfigure}[b]{0.091\textwidth}
        \centering
        \includegraphics[trim={180 50 100 0},clip,width=\textwidth]{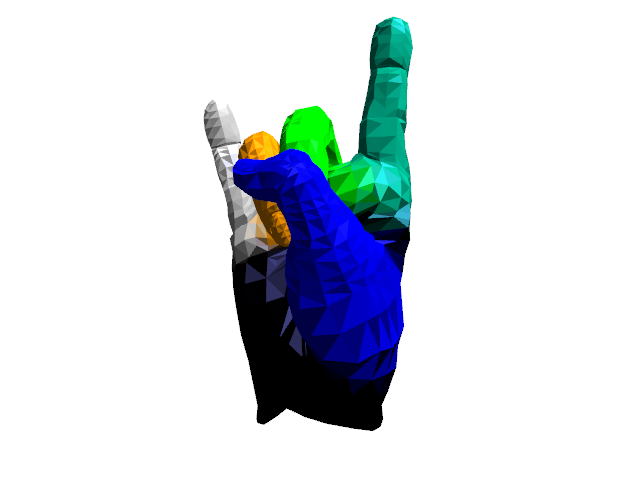}
    \end{subfigure}
    \begin{subfigure}[b]{0.091\textwidth}
        \centering
        \includegraphics[trim={110 50 140 95},clip,width=\textwidth]{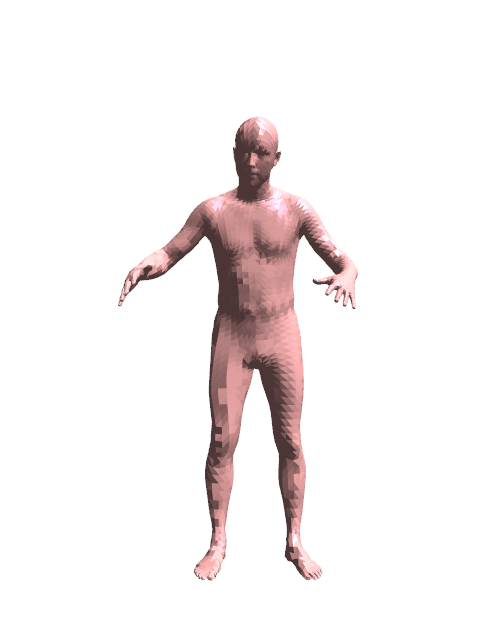}
    \end{subfigure}
    \begin{subfigure}[b]{0.091\textwidth}
        \centering
        \includegraphics[trim={110 50 140 95},clip,width=\textwidth]{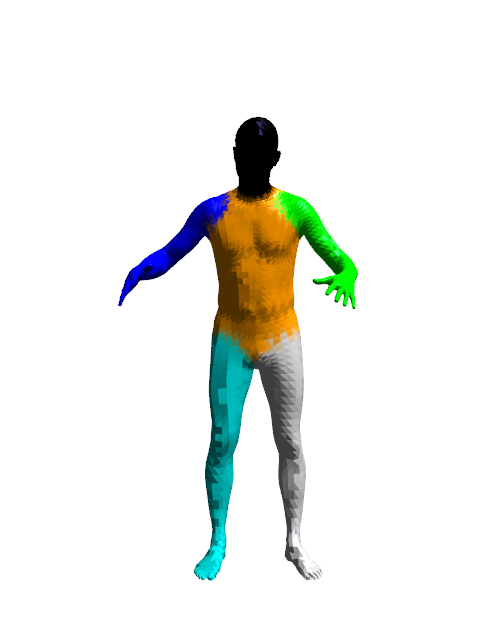}
    \end{subfigure}
    \begin{subfigure}[b]{0.091\textwidth}
        \centering
        \includegraphics[trim={110 50 140 95},clip,width=\textwidth]{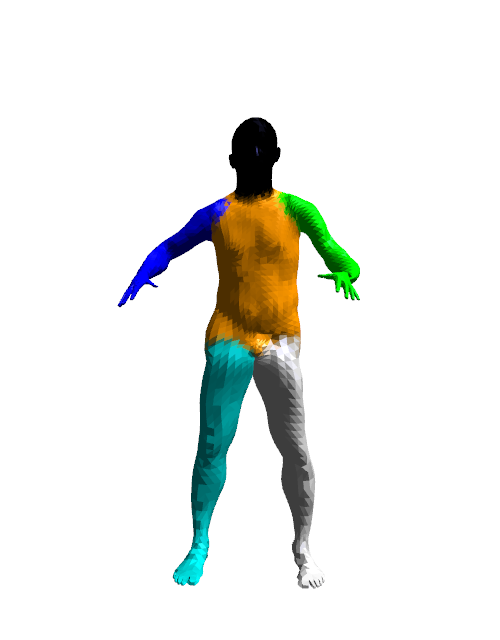}
    \end{subfigure}
    \begin{subfigure}[b]{0.091\textwidth}
        \centering
        \includegraphics[trim={110 50 140 95},clip,width=\textwidth]{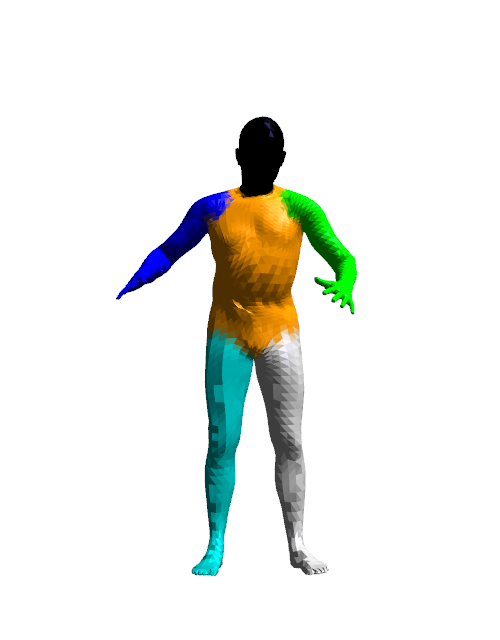}
        
    \end{subfigure}
    \begin{subfigure}[b]{0.091\textwidth}
        \centering
        \includegraphics[trim={110 50 140 95},clip,width=\textwidth]{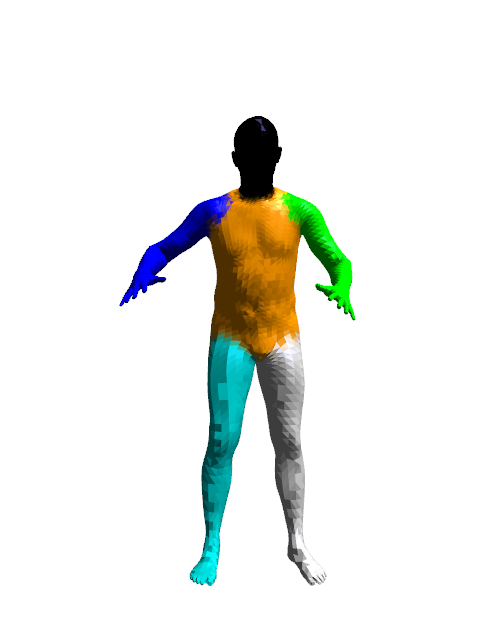}
        
    \end{subfigure}

    \begin{subfigure}[b]{0.091\textwidth}
        \centering
        \includegraphics[trim={110 50 140 65},clip,width=\textwidth]{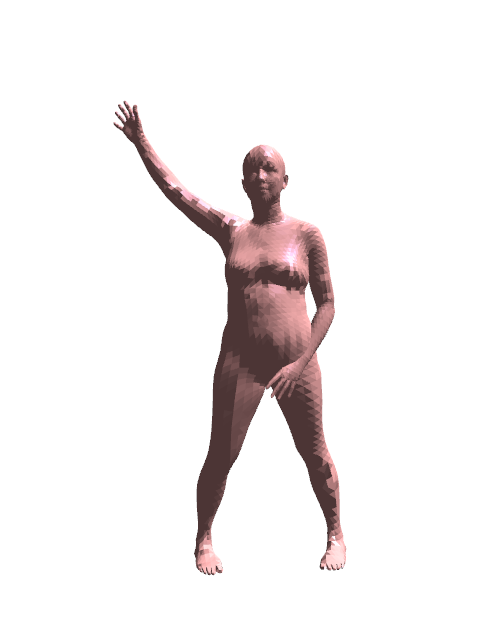}
        \caption{Input}
        
    \end{subfigure}
    \begin{subfigure}[b]{0.091\textwidth}
        \centering
        \includegraphics[trim={110 50 140 65},clip,width=\textwidth]{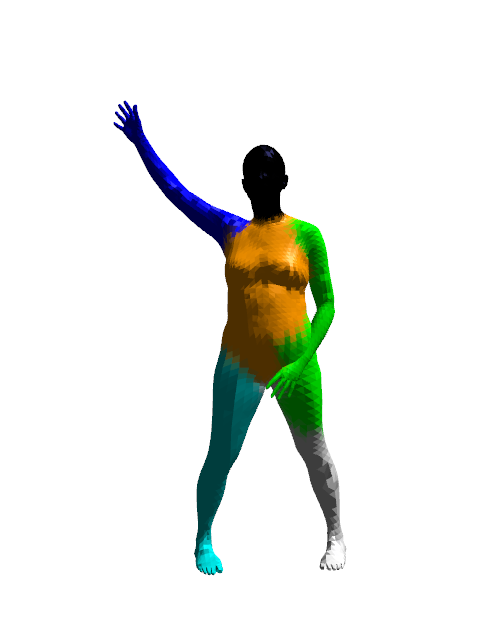}
        \caption{Segment}
        
    \end{subfigure}
    \begin{subfigure}[b]{0.091\textwidth}
        \centering
        \includegraphics[trim={110 50 140 65},clip,width=\textwidth]{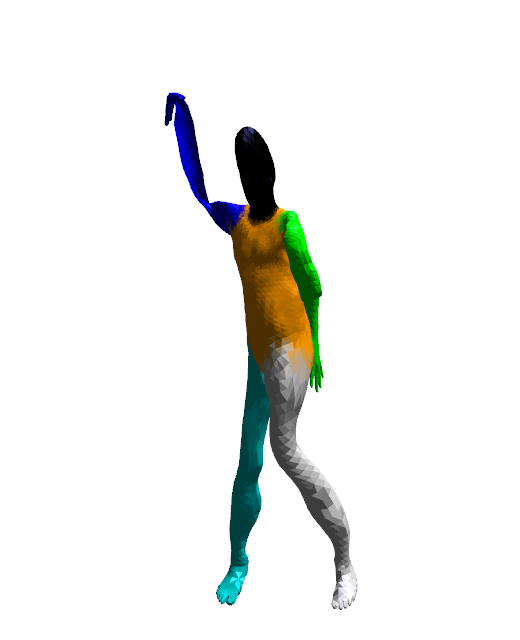}
        \caption{\cite{groueix20183d}}
        
    \end{subfigure}
    \begin{subfigure}[b]{0.091\textwidth}
        \centering
        \includegraphics[trim={110 50 140 65},clip,width=\textwidth]{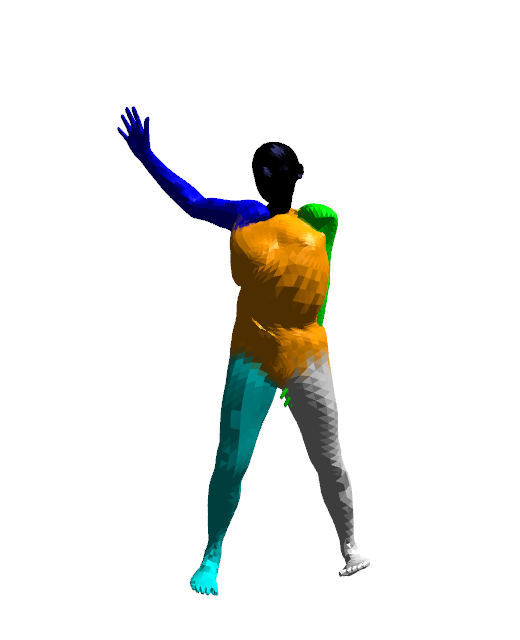}
        \caption{CD}
        
    \end{subfigure}
    \begin{subfigure}[b]{0.091\textwidth}
        \centering
        \includegraphics[trim={110 50 140 65},clip,width=\textwidth]{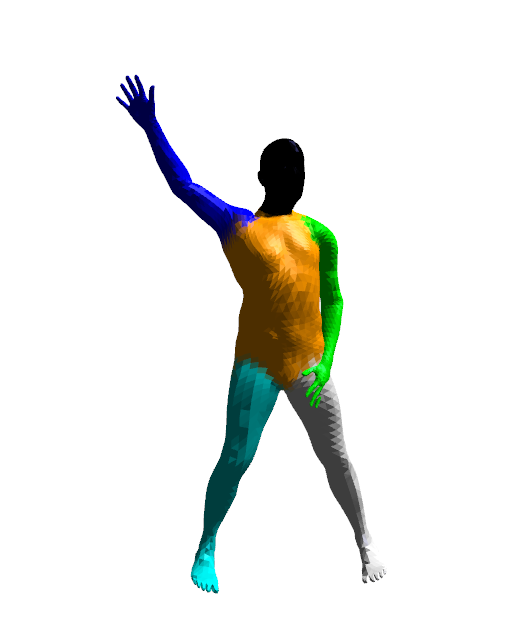}
        \caption{LBS-AE}
        
    \end{subfigure}
    \vspace{-7pt}
    \caption{Qualitative comparisons on captures hands and SURREAL (SMPL). Given  point clouds sampled from the surfaces
	of input shapes (a), (c-e) are the reconstructions from different algorithms. (b) is the inferred segmentation of
	LBS-AE on the input shape.} 
    \label{fig:qual_hand}
\end{figure}

We fit and reconstruct the hand and body data as shown in Figure~\ref{fig:qual_hand}. 
For the thumb-up pose, due to wrong correspondences from nearest neighbor search, both~\cite{groueix20183d} and
LBS-AE$_{\text{CD}}$  reconstruct wrong poses. 
The wrong correspondence causes problems to~\cite{groueix20183d}. Since the deformation from templates to targeted shapes fully relies on a deep neural network, when the correspondence 
is wrong and the network is powerful, it learns distorted deformation even with a Laplacian regularization. 
On the other hand, since LBS-AE$_{\text{CD}}$ still utilizes LBS, the deformation network $d$ is easier to regularize,
which results in better finger reconstructions. 
We note that~\cite{groueix20183d} learns proper deformation if the correspondence can be found correctly, such as 
the third row in Figure~\ref{fig:qual_hand}. In both cases, the proposed LBS-AE can learn segmentation well and recover the poses better. 

\begin{figure}
    \centering
    \begin{subfigure}[b]{0.091\textwidth}
        \centering
        \includegraphics[trim={110 100 140 95},clip,width=\textwidth]{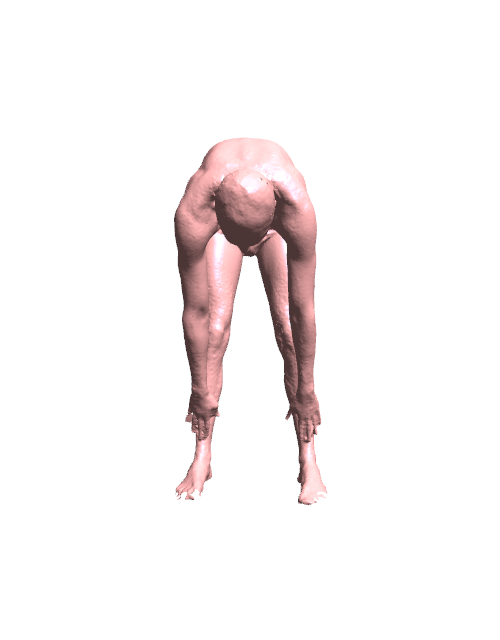}
    \end{subfigure}
	\hfill
    \begin{subfigure}[b]{0.091\textwidth}
        \centering
        \includegraphics[trim={110 100 140 95},clip,width=\textwidth]{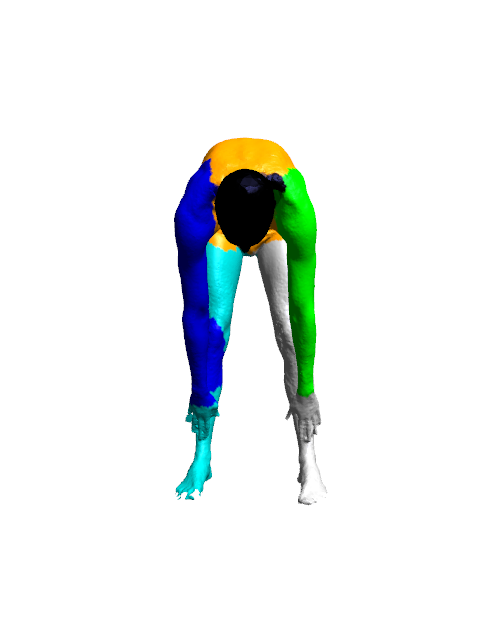}
        
    \end{subfigure}
	\hfill
    \begin{subfigure}[b]{0.091\textwidth}
        \centering
        \includegraphics[trim={110 100 140 95},clip,width=\textwidth]{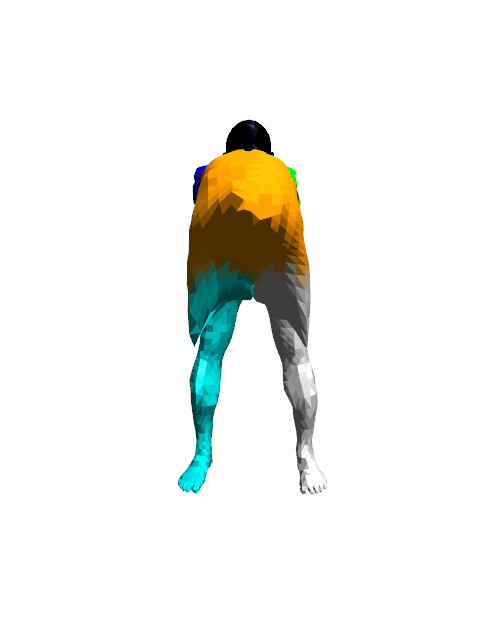}
        
    \end{subfigure}
	\hfill
    \begin{subfigure}[b]{0.091\textwidth}
        \centering
        \includegraphics[trim={110 100 140 95},clip,width=\textwidth]{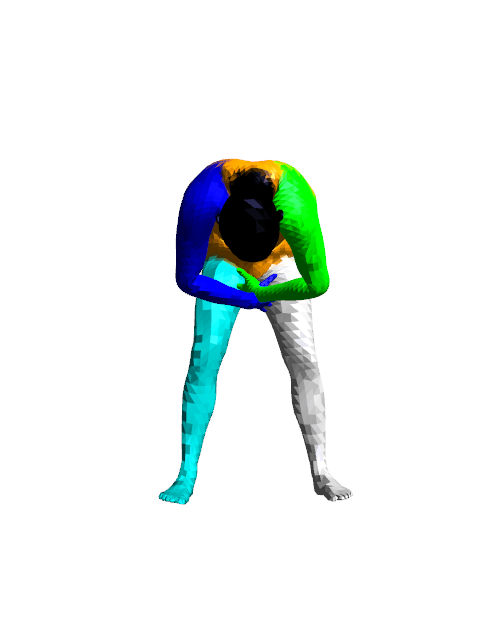}
        
    \end{subfigure}
	\hfill
    \begin{subfigure}[b]{0.091\textwidth}
        \centering
        \includegraphics[trim={110 100 140 95},clip,width=\textwidth]{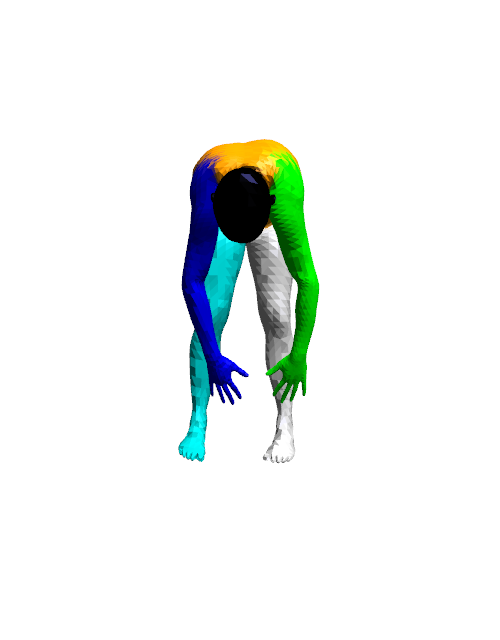}
        
    \end{subfigure}

    \begin{subfigure}[b]{0.091\textwidth}
        \centering
        \includegraphics[trim={110 30 100 80},clip,width=\textwidth]{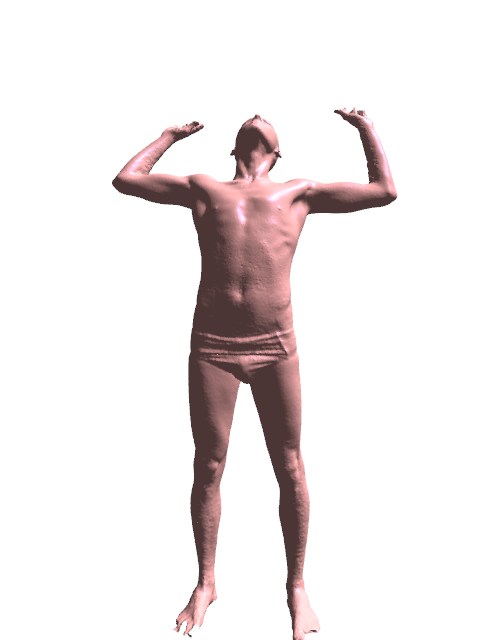}
        \caption{Input}
        
    \end{subfigure}
	\hfill
    \begin{subfigure}[b]{0.091\textwidth}
        \centering
        \includegraphics[trim={110 30 100 80},clip,width=\textwidth]{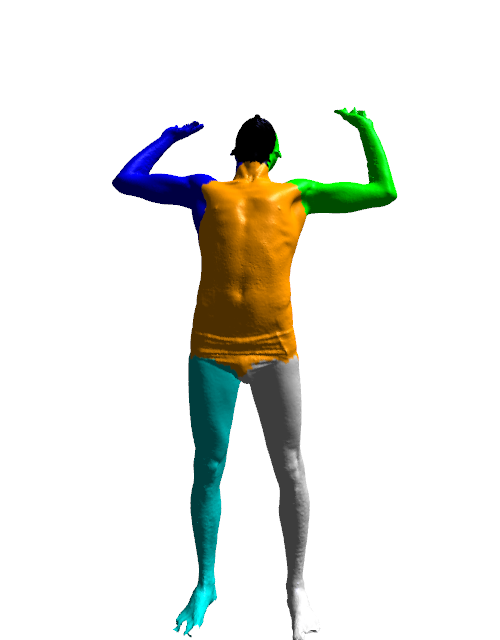}
        \caption{Segment}
        
    \end{subfigure}
	\hfill
    \begin{subfigure}[b]{0.091\textwidth}
        \centering
        \includegraphics[trim={110 30 100 80},clip,width=\textwidth]{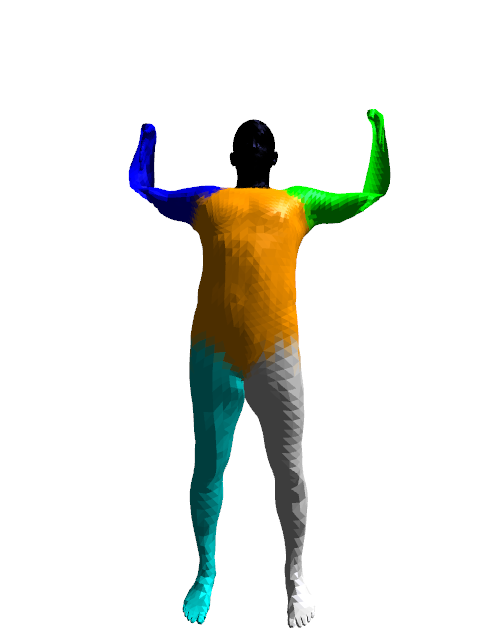}
        \caption{\cite{groueix20183d}}
        
    \end{subfigure}
	\hfill
    \begin{subfigure}[b]{0.091\textwidth}
        \centering
        \includegraphics[trim={110 30 100 80},clip,width=\textwidth]{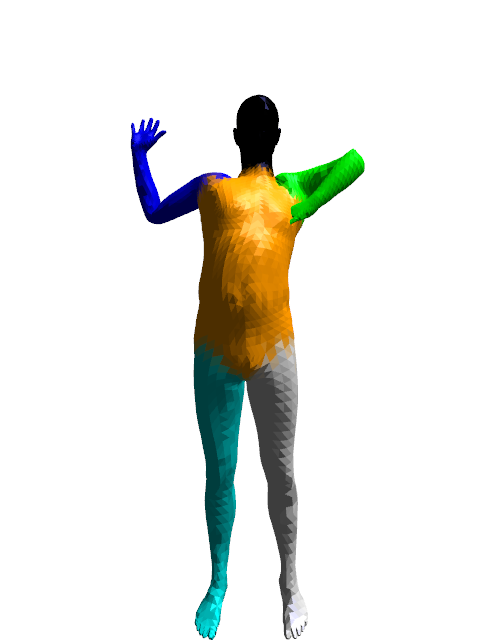}
        \caption{CD}
    \end{subfigure}
	\hfill
    \begin{subfigure}[b]{0.091\textwidth}
        \centering
        \includegraphics[trim={110 30 100 80},clip,width=\textwidth]{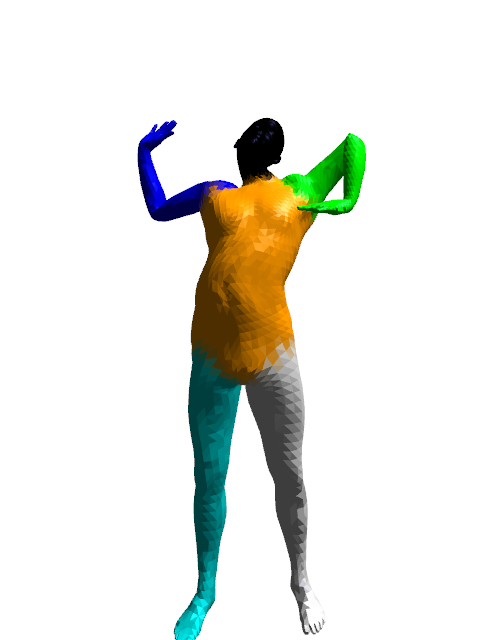}
        \caption{LBS-AE}
    \end{subfigure}
    
    \vspace{-5pt}
    \caption{Qualitative Comparison on FAUST.} 
    \vspace{-10pt}
    \label{fig:qual_faust}
\end{figure}

Lastly, we consider fitting FAUST, with only $200$ samples, as shown in Figure~\ref{fig:qual_faust}. 
With limited and diverse poses, we have less hint 
of how the poses deform~\cite{wei2016dense}, a nearest neighbor search is easily trapped in bad local optimums as we mentioned in Figure~\ref{fig:local}.
The proposed LBS-AE still results in reasonable reconstructions and segmentation, though the right arm in the second row suffers from the local optimum 
issues within the segmentation. 
A fix is to learn more fine-grained segmentation, but it brings the trade-off between task difficulty and model capacity, which we leave for future work.

\subsection{Quantitative Study}
\label{sec:quant}
We conduct quantitative analysis on reconstruction, pose estimation, and correspondence on synthetic hand and body data. 
We use $\sqrt{\text{CD}}$ as the proxy to reconstructions.
Pose estimation compares the average $\ell_2$ distance between true joint positions and inferred ones while
correspondence also measures the average $\ell_2$ between found and true correspondences.
We randomly generate $4000$ testing pairs from the testing data for correspondence comparison. 
Given two shapes, we fit the shapes via the trained models. 
Since we know the correspondence of the reconstructions, we project the data onto the reconstructions to find the correspondence. 
For more details, we refer readers to~\cite{groueix20183d}. 

We compare three variants of~\cite{groueix20183d}, including the supervised version with full correspondence, and the
unsupervised version with and without synthetic data augmentation aforementioned.
For LBS-AE, we also consider three variants, including a simple CD baseline (LBS-AE$_{\text{CD}}$), 
a segmentation network $s$  trained on poses from uniform distributions LBS-AE$_{\text{RAND}}$ and
joint training version (LBS-AE).
The results are shown in Table~\ref{tb:quant}.

For LBS-AE variants, the jointly trained LBS-AE is better than LBS-AE$_{\text{CD}}$ and LBS-AE$_{\text{RAND}}$. 
It supports the hypothesis in Section~\ref{sec:seg_exp}, that joint training facilitates improving model fitting and segmentation. 
Also, as shown in Section~\ref{sec:seg_exp}, the pretrained segmentation network still has reasonable testing accuracy and brings an improvement over using CD loss only. 
On the other hand, the supervised version of~\cite{groueix20183d} trained with full correspondence is worse than the
proposed unsupervised LBS-AE due to generalization ability.
For correspondence on the SMPL training set, supervised \cite{groueix20183d} achieves $0.065$ while LBS-AE achieve $0.069$. 
If we increase the training data size three times, supervised~\cite{groueix20183d} improves its correspondence result to be $0.095$.
For hand data, supervised~\cite{groueix20183d} generalizes even worse with 
only $1500$ training examples. It suggests that leveraging LBS models into the model can not only use smaller networks but also generalize better than relying on an unconstrained deformation from a deep network.

\begin{table}
        \small
        \centering
        \setlength{\tabcolsep}{0.4em}
    \begin{tabular}{c|c|c|c|c|c|c}
        \hline
        & \multicolumn{3}{c|}{SMPL}& \multicolumn{3}{c}{Syn. Hand} \\
        \hline
        Algorithm & Recon & Pose & Corre.& Recon & Pose & Corre. \\
        \hline
        Unsup.~\cite{groueix20183d} & 0.076 & 0.082 & 0.136 & 0.099 & 0.035 & 0.176  \\
        Unsup.+Aug~\cite{groueix20183d} & 0.081 & 0.081 & 0.132  & 0.069 & 0.049 & 0.140 \\
        Sup.~\cite{groueix20183d} & 0.073 & 0.071 & 0.104& 0.062 & 0.047 & 0.135  \\
        \hline
        LBS-AE$_{\text{CD}}$ & 0.051 & 0.152 & 0.147 & 0.082 & 0.069 & 0.168 \ \\
         LBS-AE$_{\text{RAND}}$& 0.041 & 0.058 & 0.100  & 0.069 & 0.050 & 0.137\\
        LBS-AE & {\bf 0.037} & {\bf 0.048} & {\bf 0.091} & {\bf 0.053} & {\bf 0.035} & {\bf 0.111}  \\
        \hline
    \end{tabular}
    \caption{Quantitative results on synthetic data.}
    \label{tb:quant}
\end{table}

\vspace{-8pt}
\paragraph{Deformation Network}
We also investigate the ability of the deformation in LBS-AE.  
For data generated via SMPL, we know the ground truth of deformed templates $\Ub^{gt}$ of each shape. The average $\ell_2$ distance
between corresponding points from $\Ub^{gt}$ and $\Ub^d$ is $0.02$, while the average distance between $\Ub^{gt}$ and
$\Ub$ is $0.03$.

\begin{figure}
    \centering
    \begin{subfigure}[b]{0.155\textwidth}
        \centering
        \includegraphics[trim={130 100 130 20},clip,width=0.48\textwidth]{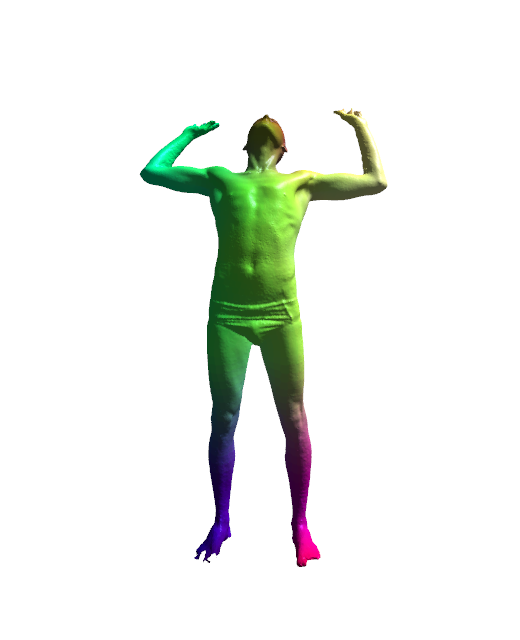}
        \includegraphics[trim={130 100 130 20},clip,width=0.48\textwidth]{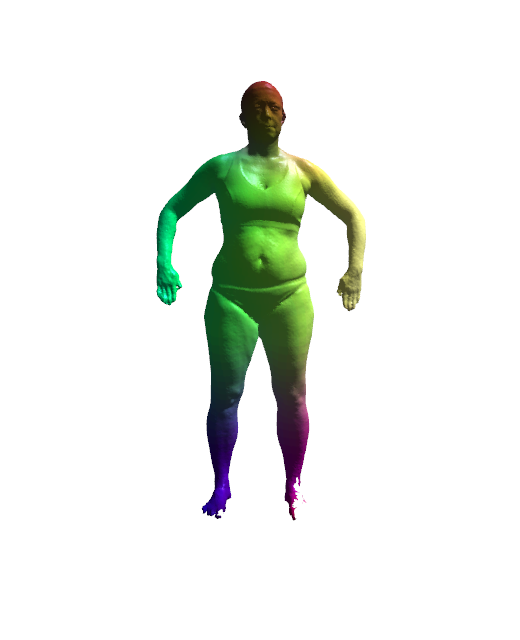}
        \caption{}
    \end{subfigure}
    \hfill
    \begin{subfigure}[b]{0.155\textwidth}
        \centering
        \includegraphics[trim={130 100 130 20},clip,width=0.48\textwidth]{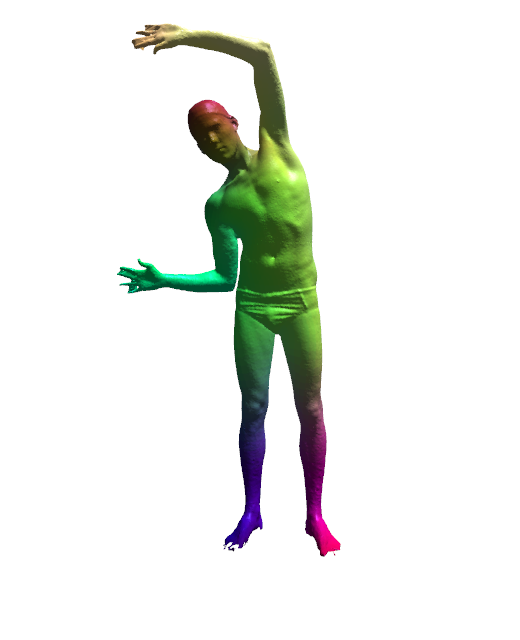}
        \includegraphics[trim={130 100 130 20},clip,width=0.48\textwidth]{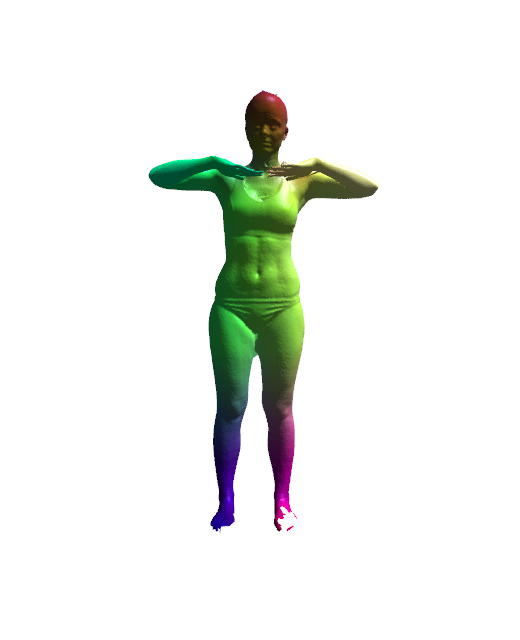}
        \caption{}
    \end{subfigure}
    \hfill
    \begin{subfigure}[b]{0.155\textwidth}
        \centering
        \includegraphics[trim={130 100 130 20},clip,width=0.48\textwidth]{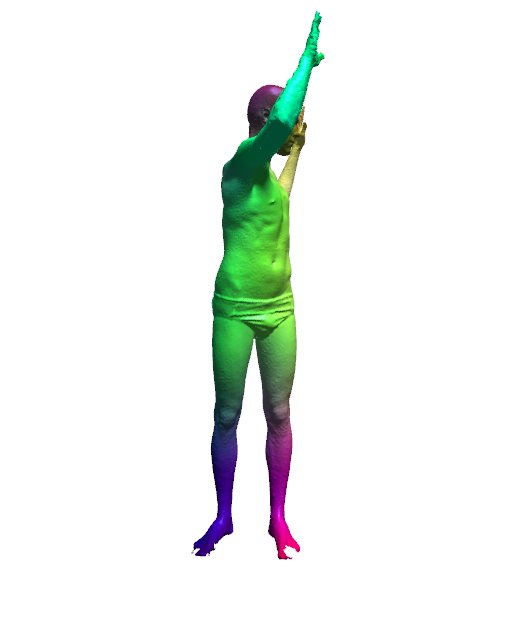}
        \includegraphics[trim={130 100 130 20},clip,width=0.48\textwidth]{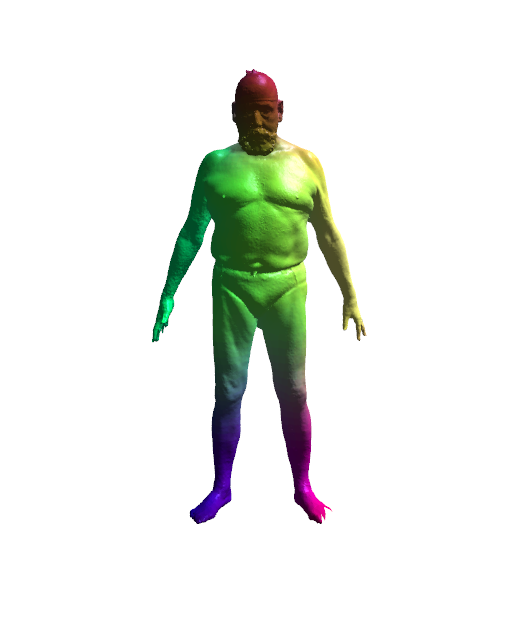}
        \caption{}
    \end{subfigure}
    \vspace{-20pt}
    \caption{Inferred correspondence of FAUST testing data.}
    \vspace{-12pt}
    \label{fig:corr}
\end{figure}

\vspace{-8pt}
\paragraph{Real-World Benchmark.}
One representative real-world benchmark is FAUST~\cite{bogo2014faust}.  
We follow the protocol used in~\cite{groueix20183d}  for comparison,
where they train on SMPL with SURREAL parameters and then fine-tune on FAUST.
In~\cite{groueix20183d}, they use a different number of data from SMPL with
SURREAL parameters, while we only use 23K. The numerical results are
shown in Table~\ref{tb:faust}. 
With only 23K SMPL data and self-supervision, we are better than unsupervised~\cite{groueix20183d} with 50K data, supervised~\cite{groueix20183d} with 10K data,  and the supervised learning algorithm FMNet~\cite{litany2017deep}.
We show some visualization of the inferred correspondence in Figure~\ref{fig:corr}.

\begin{table}
        \centering
    \begin{tabular}{c|c|c}
        \hline
        Algorithm & Inter. error (cm) & Intra. err (cm) \\
        \hline
        FMNet~\cite{litany2017deep} & 4.826 & 2.44 \\
        Unsup.~\cite{groueix20183d} (230K) & 4.88 & - \\
        Sup.~\cite{groueix20183d} (10K) & 4.70 &  - \\
        Sup.~\cite{groueix20183d} (230K) & 3.26 & 1.985 \\
        \hline
        LBS-AE (23K) & 4.08 & 2.161\\
        \hline
    \end{tabular}
    \vspace{-5pt}
    \caption{Correspondence results on FAUST testing set.}
    \vspace{-8pt}
    \label{tb:faust}
\end{table}

\section{Conclusion}
We propose a self-supervised autoencoding algorithm, LBS-AE, to align articulated mesh models to point clouds.
The decoder leverages an artist-defined mesh rig, and using LBS. We constrain the encoder to infer interpretable joint angles. 
We also propose the structured Chamfer distance for training LBS-AE, defined by inferring a meaningful segmentation of
the target data to improve the correspondence finding via nearest neighbor search in the original Chamfer distance.  
By combining LBS-AE and the segmentation inference, 
we demonstrate we can train these two components simultaneously without supervision (labeling) from data.
As training progress, the proposed model can start adapting to the data distribution and improve with self-supervision.
In addition to opening a new route to model fitting without supervision, 
the proposed algorithm also provides a successful example showing how to encode existing prior knowledge in a geometric deep learning model.

{\small
\bibliographystyle{ieee}
\bibliography{main}
}

\end{document}